\documentclass[11pt]{article}

\usepackage{graphicx}
\usepackage{subfigure}
\usepackage{multirow}
\usepackage{amssymb}
\usepackage{url}
\usepackage{amsmath}
\usepackage{amsfonts}
\usepackage{algorithmic} 
\usepackage{algorithm}

\begin{document}
\sloppy

\title{\bf Affine Image Registration Transformation Estimation Using a Real Coded Genetic Algorithm with SBX}

\author{    {\bf Mosab Bazargani}\\
            \small CENSE and DEEI-FCT\\
            \small Universidade do Algarve\\
            \small Campus de Gambelas\\
            \small 8005-139 Faro, Portugal\\
            \small mbazargani@gmail.com
\and
           {\bf Ant\'{o}nio dos Anjos}\\
           \small Departamento de Ci\^{e}ncias e Tecnologias\\
           \small ISMAT - Instituto Superior Manuel Teixeira Gomes\\
           \small 8500-508 Portim\~{a}o, Portugal\\
           \small antoniodosanjos@gmail.com
\and
            {\bf Fernando G. Lobo}\\
            \small CENSE and DEEI-FCT\\
            \small Universidade do Algarve\\
            \small Campus de Gambelas\\
            \small 8005-139 Faro, Portugal\\
            \small fernando.lobo@gmail.com  
\and
            {\bf Ali Mollahosseini}\\
            \small ILab 2.57, DEEI-FCT\\
            \small Universidade do Algarve\\
            \small Campus de Gambelas\\
            \small 8005-139 Faro, Portugal\\
            \small ali.mollahosseini@gmail.com
\and
            {\bf Hamid Reza Shahbazkia}\\
            \small ILab 2.57, DEEI-FCT\\
            \small Universidade do Algarve\\
            \small Campus de Gambelas\\
            \small 8005-139 Faro, Portugal\\
            \small hshah@ualg.pt
}
\date{}
\maketitle

\begin{abstract}
This paper describes the application of a real coded genetic algorithm (GA) to align two or more 2-D images by means of image registration. The proposed search strategy is a transformation parameters-based approach involving the affine transform. The real coded GA uses Simulated Binary Crossover (SBX), a parent-centric recombination operator that has shown to deliver a good performance in many optimization problems in the continuous domain. In addition, we propose a new technique for matching points between a warped and static images by using a randomized ordering when visiting the points during the matching procedure. This new technique makes the evaluation of the objective function somewhat noisy, but GAs and other population-based search algorithms have been shown to cope well with noisy fitness evaluations. The results obtained are competitive to those obtained by state-of-the-art classical methods in image registration, confirming the usefulness of the proposed noisy objective function and the suitability of SBX as a recombination operator for this type of problem.
\end{abstract}

Image Registration (IR) is the process of aligning two or more images of the same scene taken at different times, from different directions, and/or by different sensors, by finding the best mapping function between them~\cite{Brown:1992,ZitovaFlusser:2003}. Most research in this area is based on classical algorithms and methods, but during the past two decades or so, there has been a growing interest in the application of Evolutionary Computation (EC) and other Metaheuristic (MH) methods to solve the problem.

This paper addresses the design of the mapping function for 2-D image registration using the affine transform. In this problem, the goal is to find the best mapping function, also called transform, that warps a \emph{Deformed} image (\emph{D}) in the direction of a \emph{Static} image (\emph{S}), based in the images' features (e.g. point positions). The objective of the registration procedure is to find the optimum, or an \emph{acceptable} sub-optimal, mapping function.

To solve this problem we use an elitist real coded genetic algorithm using Simulated Binary Crossover and Gaussian mutation. The results obtained are superior to those obtained by a previous real coded genetic algorithm on the same synthetic images, and are also very competitive with classical state-of-the-art image registration algorithms.

The paper is organized as follows. Section~\ref{sec:IR} describes the IR problem and the two broad categories of methods commonly used. It also describes the affine transformation, a commonly used linear transformation whose parameters correspond to the decision variables of the optimization problem that is formulated. Section~\ref{sec:RW} presents a brief review of related work that has been proposed to address the IR problem, both with classical as well as with EC and MH methods. Section~\ref{sec:defineproblem} presents a real coded GA formulation for the IR problem. Section~\ref{sec:ER} presents and discusses experimental results. Finally, Section~\ref{sec:conclusion} concludes the paper.

\section{Image Registration}
\label{sec:IR}
Image registration has been applied in a large number of research areas, including medical image analysis, computer vision and pattern recognition~\cite{ZitovaFlusser:2003}. In all IR problems there are at least two images, a \emph{Deformed} (\emph{D}) and a \emph{Static} (\emph{S}) image, that represent the same object, or scene. IR aims to find the best mapping function \emph{T} to warp \emph{D} towards \emph{S}, as shown below,
\begin{equation}
\label{eq:warpped}
   W \cong T(D) \enspace .
\end{equation}
\emph{W} is a warped image which should be as closely shaped to \emph{S} as possible.
IR typically has the four following steps \cite{ZitovaFlusser:2003}:
\begin{enumerate}
 \item Feature detection;
 \item Feature matching;
 \item Mapping function design;
 \item Image transformation and re-sampling.
\end{enumerate}
In order to find the correct transformation, image features such as closed-boundary regions, edges, line intersections, corners, and so on, should be extracted (feature detection). These features can be used as control points. Correspondences have to be set between the extracted features of \emph{S} and \emph{D} (feature matching). Then, the type of transforming model has to be defined and its parameters estimated (mapping function). Finally, the D image is transformed by means of the mapping function (image transformation).

IR can be seen as a function approximation method~\cite{SeixasOchiConciSaade:2008}, and it is an NP-Complete problem~\cite{KeysersUnger:2003}. The most important aspect of IR is the discovery of the unknown parametric transformation that relates the two images. Two different approaches can be found in the literature:

\begin{itemize} 
\item{Matching-based approaches}
\item{Transformation parameters-based approaches}
\end{itemize} 

\emph{Matching-based} approaches conduct a search within the space of possible feature correspondences (typically point matching) between the two images. Thereafter, a registration transformation is obtained based on the correspondence found.
In contrast, \emph{transformation parameters-based} approaches perform a direct search in the space of the parameters of the transformation. 

IR methods can also be classified according to the type of models that they allow to transform the \emph{D} image into the \emph{S} image (sometimes also referred in the literature as the \emph{scene} and \emph{model} images). Two major types of models are used: \emph{linear} and \emph{non-linear transformations}. Linear transformations preserve the operations of vector addition and scalar multiplication. The same does not hold for non-linear (or \emph{elastic}) transformations, which allow local deformations of the image.  
Although, with a few modifications, the proposed approach could be adapted to work with polynomial transforms (e.g. quadratic and cubic), in this paper we only deal with linear transformation models. Several such models exist and one of the most general is the \emph{affine transformation} which is discussed next.

\subsection{Affine Transform}
\label{sec:IR.AT}
\begin{figure}[!ht]
  \centering
      \includegraphics{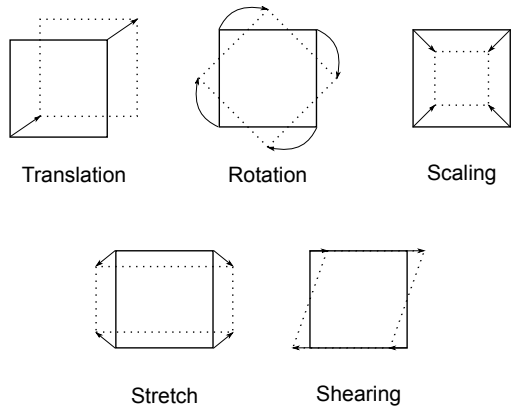}
  \caption{Elementary geometric transforms for a planar surface element used in the affine transform: translation, rotation, scaling, stretching, and shearing.}
  \label{fig:fig1}
\end{figure}

The affine transform is a linear transformation that includes the following elementary transformations: translation, rotation,
scaling, stretching, and shearing~\cite{Jahne:2005}. These elementary transformations are illustrated in Figure~\ref{fig:fig1}. 

A geometric operation transforms a given image \emph{D} into a new image \emph{W} by modifying the coordinates
of the image points, as follows:
\begin{equation}
\label{eq:warpping}
	D(x,y)\stackrel{T}{\longrightarrow} W(x',y')\enspace.
\end{equation}
The original values of image \emph{D} which are located at \emph{$(x,y)$} warp to the new position \emph{$(x',y')$} in the new image 
\emph{W}. To model this process, we first need a mapping function \emph{$T$} that is a continuous coordinate transform.
An affine transformation function works in the 2-D space, thus, the search space is:
\begin{equation}
\label{eq:searchspace}
	T: \mathbb{R}^2 \Longrightarrow \mathbb{R}^2\enspace.
\end{equation}
The mapping function can be redefined as:
\begin{equation}
\begin{array}{c}
	W = T(D)\\
	\quad T: \mathbb{R}^2\Longrightarrow\mathbb{R}^2\enspace. 
\end{array}
\end{equation}
The warped image ${W(x',y')}$ is often specified as the following two separated functions for the \emph{x} and \emph{y} components:
\begin{eqnarray}
\label{eq:warped_x'_y'}
	&x' = T_x(x,y)\\
	&\quad y' = T_y(x,y)\enspace.
\end{eqnarray}
An affine transformation can be expressed by vector addition and matrix multiplication as shown in Equation~\ref{eq:affinetrans1},
\begin{equation}
\label{eq:affinetrans1}
	\left[
	\begin{array}{c}
  	x'\\
  	y'
	\end{array} \right]
	= S\left[
  	\begin{array}{lr}
    	\cos\theta & -\sin\theta\\
    	\sin\theta & \cos\theta
  	\end{array} \right]
 	\left[
	\begin{array}{c}
  	x\\
  	y
 	\end{array} \right] +
 	\left[
	\begin{array}{c}
  	t_x\\
  	t_y
	\end{array} \right]
\end{equation}
where $S$ is the scaling parameter. By multiplying $S$ with the rotation matrix, Equation \ref{eq:affinetrans1} can be written as:
\begin{equation}
\label{eq:affinetrans2}
	\left[
	\begin{array}{c}
  	x'\\
  	y'
	\end{array} \right]
	= \left[
  	\begin{array}{lr}
    	a_{11} & a_{12}\\
    	a_{21} & a_{22}
  	\end{array} \right]
	\left[
	\begin{array}{c}
  	x\\
  	y
 	\end{array} \right] +
	\left[
	\begin{array}{c}
  	t_x\\
  	t_y
	\end{array} \right]\enspace.
\end{equation}
Finally, by using homogeneous coordinates, the affine transformation can be rewritten as Equation~\ref{eq:affinetrans3}.
\begin{equation}
\label{eq:affinetrans3}
	\left[
	\begin{array}{c}
	x'\\
	y'\\
	1
	\end{array} \right]
	= \left[
  	\begin{array}{ccc}
    	\theta_{0} & \theta_{1} & \theta_{2}\\
    	\theta_{3} & \theta_{4} & \theta_{5}\\
    	0 & 0 & 1
  	\end{array} \right]
	\left[
	\begin{array}{c}
  	x\\
  	y\\
  	1
	\end{array} \right]\enspace.
\end{equation}
The affine transform has six parameters: $\mathit{\theta_0}$, $\mathit{\theta_1}$, $\mathit{\theta_2}$, $\mathit{\theta_3}$, $\mathit{\theta_4}$, and $\mathit{\theta_5}$. $\mathit{\theta_2}$ and $\mathit{\theta_5}$ specify
the translation and $\mathit{\theta_{0}}$, $\mathit{\theta_{1}}$, $\mathit{\theta_{3}}$, and $\mathit{\theta_{4}}$ aggregate rotation, scaling, stretching, and shearing.

\section{Related Work}
\label{sec:RW}

There is variety of techniques for solving IR problems. 
This  section presents a brief review of some of the most important ones, including both classical as well as evolutionary computation and metaheuristic based approaches. 

\subsection{Classical Methods}
Most of the classical approaches for IR can be found in~\cite{ZitovaFlusser:2003, Brown:1992, ChuiRangarajan:2003}.
Two state-of-the-art approaches from this category of methods are Robust Point Matching (TPS-RPM)~\cite{Chui:2001}, and Shape Context (SC)~\cite{Belongie:2002}.

TPS-RPM is a  method for matching two point-sets in a Deterministic Annealing (DA) setting. It uses a fuzzy-like matrix instead of a binary permutation matrix to find the matching between two sets of points. In TPS-RPM, both the point correspondences and the transformations are computed interchangeably. Therefore, RPM can be viewed as a general framework for point matching and can accept different transformation models~\cite{Chui:2001} like affine, and even more complicated models like Thin Plate Splines (TPS)~\cite{Duchon:1977}. This method is a kind of a hybrid in the sense that it can be considered both a matching-based and a transformation-based approach for IR.

Shape Context (SC) is a matching-based approach that is usually used to estimate the transformation between two images, by finding matches between samples from the edges of the objects in the images. It basically consists of analyzing the spacial relationship between points. It uses four main parameters. The first defines the number of radial bins for the creation of the histograms, the second is the number of theta bins that defines how many slices should the histograms be divided into, and the third and fourth parameters, the minimum and maximum width of the bins. For more information on these and other classical IR methods, the reader is directed to~\cite{Brown:1992,ChuiRangarajan:2003,ZitovaFlusser:2003}.

\subsection{EC and MH methods}
Evolutionary Algorithms (EAs) and other metaheuristics (MHs) have been applied to solve IR problems. 
EAs and MHs are stochastic optimization methods whose goal is to find a solution or a set of solutions that perform(s) best with respect to a certain objective(s). During the last decades these algorithms have been successful in solving a variety of search and optimization problems, and the domain of image registration has been no exception. As opposed to the classical methods, which are typically based on gradient-based search, EAs and MHs tend to escape more easily from local optima and can be considered, in general, robust methods.

The first known application of evolutionary computation to image registration is due to Fitzpatrick et al.~\cite{Fitzpatrick:1984} who applied a genetic algorithm (GA) to relate angiographic images. For the subsequent 15 years or so, other EC approaches have been proposed by different authors, but most of them were based on the canonical GA with proportionate selection and a binary representation for solutions. Such a GA has severe limitations when solving optimization problems in the continuous domain, especially due to the problem of Hamming cliffs originated from the discretization of real valued variables into binary coded values, to the fixed precision that depends on the number of bits used for each decision variable, and for imposing lower and upper bounds for a variable's value. Moreover, it is known for several years that fitness proportionate selection methods have several drawbacks when compared to ordinal-based selection methods such as ranking, tournament, or truncation selection~\cite{Goldberg:1991}. Nonetheless, most of the early EC approaches for IR used such kind of GA setup~\cite{Tsang:1995, Tsang:1997, Garai:2002, Yamany:1999, Zhang:2002}. Another limitation of the early approaches was that they only dealt with translation and rotation~\cite{Fitzpatrick:1984,Garai:2002, He:2002, Yamany:1999}, ignoring scaling, stretching, and shearing.

Most modern EC applications to IR use a direct real coded representation of solutions~\cite{Rouet:2000,He:2002, GarciaVegaAguirreZaletaCoello:2002, Chow:2004, Cordon:2003, Winter:2008,SeixasOchiConciSaade:2008}. Besides EC, other MH approaches have been applied to IR, namely Tabu Search~\cite{Wachowiak:2001}, Particle Swarm Optimization~\cite{Wachowiak:2004}, Iterated Local Search~\cite{Cordon:2006}, and Scatter Search~\cite{Cordon:2005, Santamaria:2009} to name a few. A detailed review of these works cannot be made in the paper, but the interested reader can consult recent surveys on the topic~\cite{Damas:2011,Santamaria:2011}.

\section{A real coded GA with SBX for Image Registration}
\label{sec:defineproblem}
This section introduces a real coded genetic algorithm for the optimization of the parameters of an affine transformation for the case of 2-D images. The proposed algorithm is a transformation parameters-based approach, since we are performing a direct search for the parameters that define the registration transformation. For the sake of simplicity, we assume we have two 2-D synthetic point-sets representing features from the two images. In other words, it is assumed that the feature detection step of the IR pipeline has been solved beforehand.
In the next subsections, we describe the representation, the operators, and the objective function that was used in this study.

\subsection{Representation and Operators}
\label{subsec:representation-and-operators}
The representation is straightforward. For the 2-D case, the affine transformation is defined by six parameters,
$\mathit{\theta_0} \ldots \mathit{\theta_5}$, as explained in Section~\ref{sec:IR.AT}.
A candidate solution for the GA is therefore represented by a chromosome vector with six genes, each a real number.

With respect to the GA variation operators, we use Simulated Binary Crossover (SBX) proposed by Deb and Agrawal~\cite{DebAgrawal:1995} and Gaussian mutation. The utilization of SBX crossover and Gaussian mutation is a natural choice because the problem has a continuous search space. SBX uses a probability distribution to create the offspring, and it does so by biasing the offspring to be created near the parents.
SBX is a \emph{parent-centric} recombination operator because the offspring it produces are located around the parents. This behaviour contrast with \emph{mean-centric} recombination operators whose offspring are located at the center of mass of parents. Examples of the latter category are the unimodal normal distribution crossover (UNDX), simplex crossover (SPX), and blend crossover (BLX). 

It has been shown that \emph{parent-centric} operators have in general a better performance than \emph{mean-centric} operators~\cite{Deb:2002}. This has motivated our choice of SBX as a crossover operator. Surprisingly, none of the real coded GAs proposed in the literature for addressing the IR problem has used SBX or other parent-centric crossover operators. Instead most used recombination operators such as uniform~\cite{Rouet:2000, Chow:2004}, arithmetic~\cite{He:2002} and blend crossover~\cite{Cordon:2003}.

\subsection{Objective function}
\label{subsec:objective-function}

In order to guide the search for an appropriate set of parameters for the affine transformation, we need to measure the proximity between the static and the warped image (the deformed image after the affine transformation is performed). The closer the two images are, the better the affine transformation is. Since each image is represented by a point-set, we need a way to find the similarity between two point-sets. To do so we first find a correspondence between points in the warped and static images. Once the correspondence is obtained, the objective function value is the weighted similarity of the two point-sets using the Euclidean distance of the matched points.

Algorithm~\ref{alg:objective-function} gives details of the steps involved in evaluating a candidate solution. In the algorithm, upper case letters denote matrices or vectors, and lower case letters with subscripts denote a specific element of the matrix or vector. The next paragraphs  describe the major steps of Algorithm~\ref{alg:objective-function}.

\begin{algorithm}
\small
\renewcommand{\algorithmicrequire}{\textbf{Input:}}
\renewcommand{\algorithmicensure}{\textbf{Output:}}
\caption{Objective function}
\label{alg:objective-function}
\begin{algorithmic}[1]
\REQUIRE $S, D, C$\\ 
        \qquad/* $S$ is the static image points */\\
        \qquad/* $D$ is the deformed image points */\\
        \qquad/* $C$ is a chromosome */
\ENSURE Objective function value
\STATE $W\gets T(D,C)$;  
\STATE /* Euclidean distance between the point-sets */
\FORALL{$i\in \{1,\ldots,n\}$}  
	\FORALL{$j\in \{1,\ldots,k\}$}
		\STATE $\delta_{ij}\gets \lVert w_i, s_j\rVert$;
	\ENDFOR
\ENDFOR
\STATE /* Initialize correspondence matrices */
\FORALL{$i\in \{1,\ldots,n\}$}
	\FORALL{$j\in \{1,\ldots,k\}$}
		\STATE $m^\prime_{ij}\gets 0$;
		\STATE $m^{\prime\prime}_{ij}\gets 0$;
	\ENDFOR
\ENDFOR
\STATE /* Find the closest non-assigned point */
\STATE /* $O$ is the matched-order vector */
\FORALL{$i\in O$}
	\STATE $j\gets\mbox{W2S($\Delta, M^\prime, i$)}$; \qquad// $(W \rightarrow S)$
	\STATE $M^\prime_{ij}\gets 1$;
\ENDFOR
\FORALL{$j\in O$}
	\STATE $i\gets\mbox{S2W($\Delta, M^{\prime\prime}, j$)}$; \qquad// $(S \rightarrow W)$
	\STATE $M^{\prime\prime}_{ij}\gets 1$;
\ENDFOR
\STATE $Q = M^\prime + M^{\prime\prime}$;
\STATE /* Calculate weights */
\FORALL{$i\in \{1,\ldots,n\}$}
	\FORALL{$j\in \{1,\ldots,k\}$}
		\IF{$q_{ij} \neq 0$}
			\STATE $q_{ij}^*\gets q_{ij}^{-1}$;
		\ELSE
			\STATE $q_{ij}^*\gets 0$;
		\ENDIF
	\ENDFOR
\ENDFOR
\STATE /* Weighting matches */
\FORALL{$i\in \{1,\ldots,n\}$}
	\FORALL{$j\in \{1,\ldots,k\}$}
		\STATE $m_{ij}\gets m^\prime_{ij}\times q^*_{ij}$;
	\ENDFOR
\ENDFOR
\STATE $fitness\gets 0$;
\FORALL{$i\in \{1,\ldots,n\}$}
	\FORALL{$j\in \{1,\ldots,k\}$}
		\STATE $fitness \gets fitness+(m_{ij}\times \delta_{ij})$;
	\ENDFOR
\ENDFOR
\RETURN $fitness$;
\end{algorithmic}
\end{algorithm}

To compute the objective function value of a candidate solution, we start by warping the deformed image $D$ according to the parameters of the affine transformation specified in the candidate solution $C$, yielding a new point-set $W$ (line 1 of Algorithm~\ref{alg:objective-function}). Then the matching of points in $W$ into points in $S$ is modeled using a correspondence binary matrix $\mathit{M{(n,k)}}$ based on the closest-point rule, where \emph{n} and \emph{k} correspond to the number of points in the warped and static images, respectively. The closest-point is measured using the Euclidean distance between matched points. Each point in any set corresponds, at most, to one point in the other set. To find the correspondence for each point, the closest point in the other set is chosen. If the nearest point has already been assigned to another point, the next non-assigned nearest point is chosen. This procedure is performed once for finding the correspondence matrix $M'(n,k)$ from the warped set to the static set ($W\to S$), and another time for finding the correspondence matrix $M''{(n,k)}$ from the static set to the warped set ($S\to W$). This is achieved by lines 3--24 of Algorithm~\ref{alg:objective-function}.

The order in which the correspondence points are found (lines 17 and 21 of Algorithm~\ref{alg:objective-function}), plays a vital role in the resulting correspondence matrix. A match-order vector is proposed to specify the order in which the points of a given set are visited when finding its closest-point match from the other set ($W\to S$ and $S\to W$). For different orderings, different correspondences may be found. Therefore, the match-order vector is randomly created in each generation. This makes the evaluation of a candidate solution a somewhat noisy process. In a given generation, two identical solutions obtain the same objective function value. But the same thing is not necessarily true for two identical solutions from different generations. 
Fortunately, GAs are well known for being able to handle well noisy fitness evaluations due to processing a population of solutions.  
Note that we could have used a fixed pre-determined ordering for all evaluations but decided not to do so because the point matching procedure would be somewhat biased with respect to the used ordering. 

At this point (line 24 of Algorithm~\ref{alg:objective-function}), matrices $M'$ and $M''$ specify the point-matchings from $W\to S$ and $S\to W$, respectively. We then obtain matrices $Q$ and $Q^*$. $Q$ is simply the sum of $M'$ and $M''$, thus each $q_{ij}$ can have a value of 2, 1, or 0, depending on whether point $i$ matches point $j$ in both directions, in a single direction, or has no match at all.  
Matrix $Q^*$ is obtained from $Q$ by inverting the non-zero elements. Finally, matrix $M$ is calculated from the element-wise multiplication of matrices $M'$ and $Q^*$. 
The objective function is based on the weighed similarity of two point-sets using the Euclidean distance of the matched points. The points that are connected exclusively from one direction (either $W\to S$ or $S\to W$) are penalized, and those that are connected in both directions are given half weight in terms of Euclidean distance. In other words, if the connection exists in both directions the objective function value decreases. 

This objective function is very similar to the one used by Seixas et al.~\cite{SeixasOchiConciSaade:2008}. The main difference is the use of a newly generated matched order vector in each generation, which makes the point-matching procedure less dependent on a fixed ordering of visiting the points.

\section{Experimental Results}
\label{sec:ER}
This section describes the experimental results from testing the proposed GA formulation. Five point-sets available at \texttt{\url{http://noodle.med.yale.edu/~chui/rpm/TPS-RPM.zip}} are used. Each set is composed at most by 105 points. They include the deformed and static points' locations. The deformed points were generated from the static ones by a non-affined (i.e. free-form) transform. This means that it will not be possible to obtain a perfect matching of the images by using an affine transformation model alone. 
All point coordinate values have a precision of 16 floating point. 

The GA setup was the same for all data sets. Most parameter settings were tuned beforehand, and held fixed for all the experiments. We use tournament selection without replacement of size 5, SBX crossover with distribution index 2~\cite{DebAgrawal:1995}, and Gaussian mutation for all the genes. 
The crossover probability was set to 1.0 and each gene undergoes SBX with probability 0.5. 
For replacement we use a \emph{replace worst} strategy, with the worst half of the individuals of the current population being replaced by the best half of the newly generated solutions. This replacement strategy makes the GA elitist, never losing the best solution found so far. The GA ran for 500 iterations. The experiments were performed with populations of size 30, 60, 120, 240, and 480 individuals, and for each size, 100 independent runs were executed.  

In accordance to population sizing theory of GAs \cite{Harik:99b}, larger populations sizes tend to produce a better solution quality, but also at the expense of more processing time. Figure~\ref{fig:non-affine} shows the objective function value of the best individual in the population at every generation, averaged over the 100 runs, for the various populations sizes and for the various point-sets. The performance behavior is more or less identical for all images. We can observe a substantial progress for the first 50 generations, still some progress between generations 50--200, and from there on the improvements are minor. As expected, larger population sizes gives better solution quality but the improvements are negligible for populations sizes larger than 120.

\begin{figure}
  \centering
  \subfigure{
   \includegraphics [width=0.3\columnwidth]{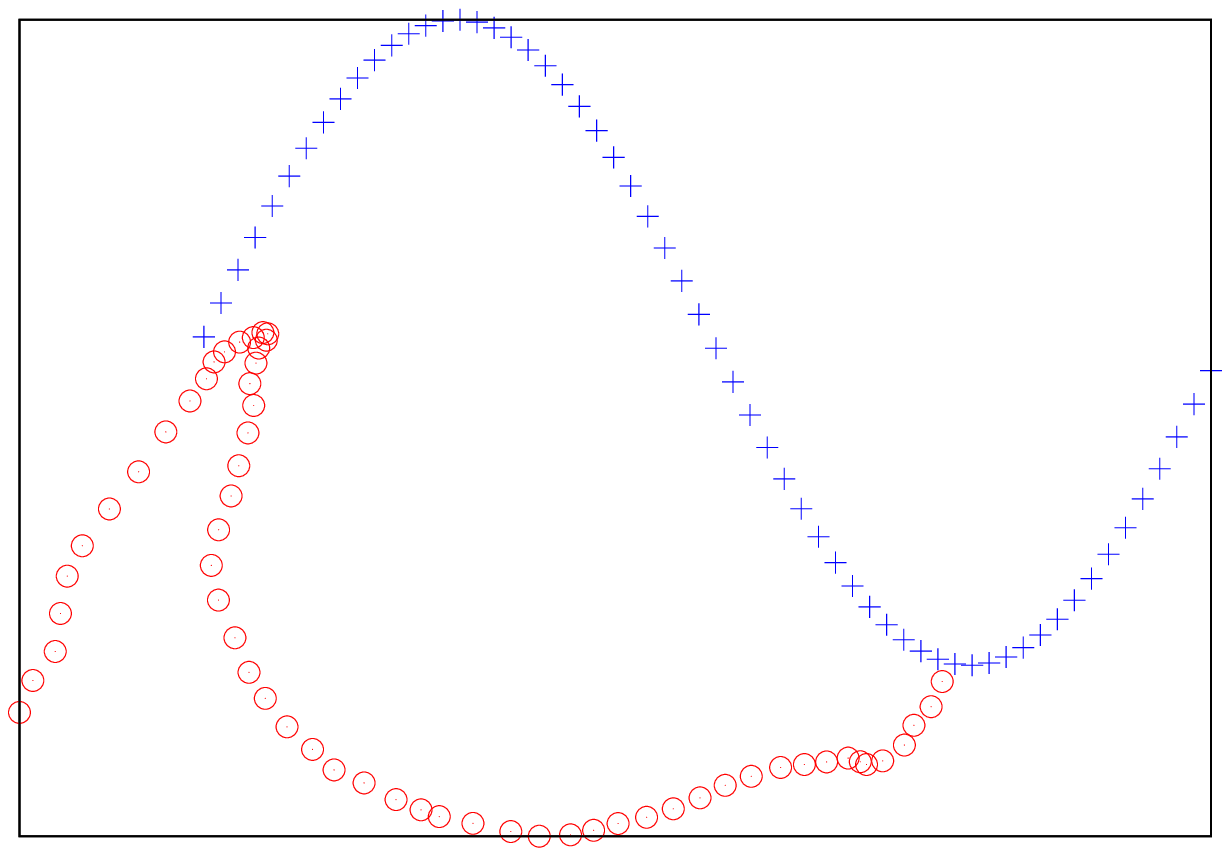}
 }
 \subfigure{
   \includegraphics [width=0.3\columnwidth]{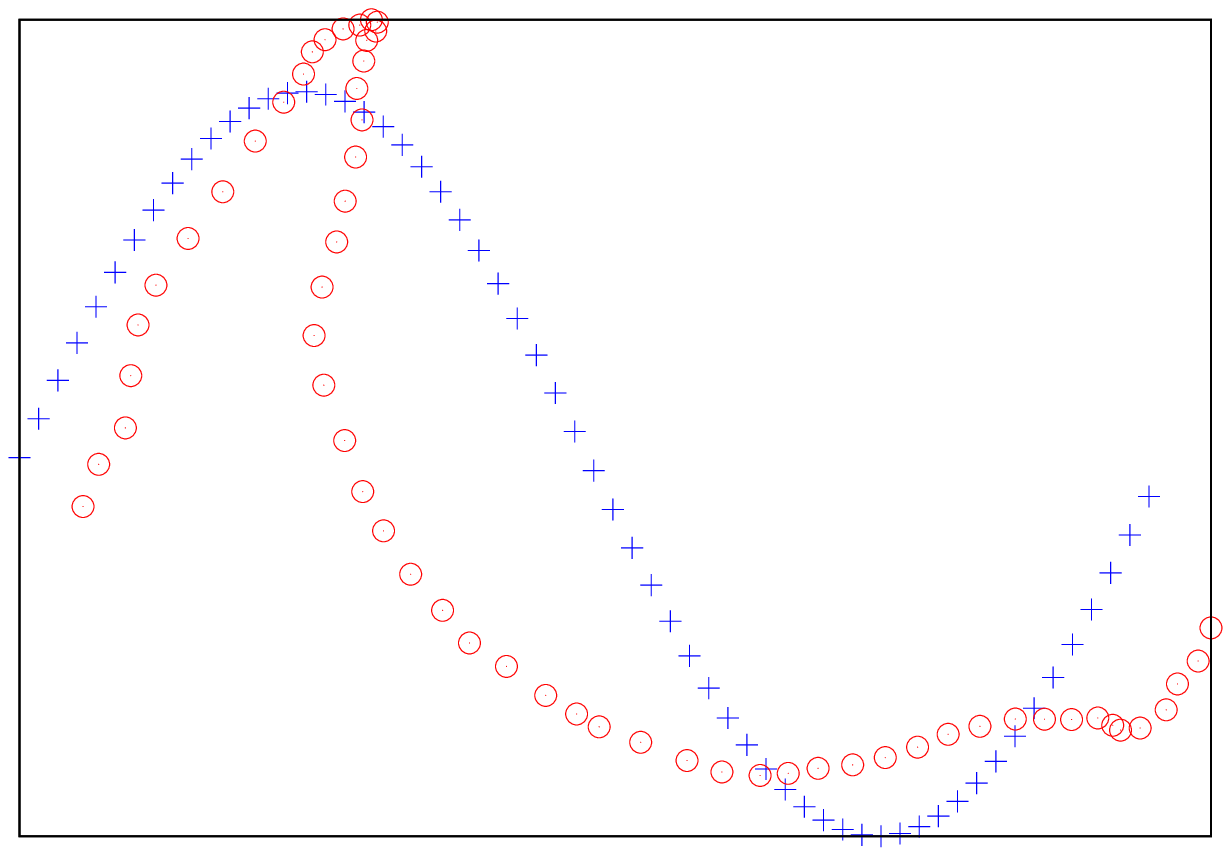}
 }\\{(a) point-set 1}\\
 \subfigure{
   \includegraphics [width=0.3\columnwidth]{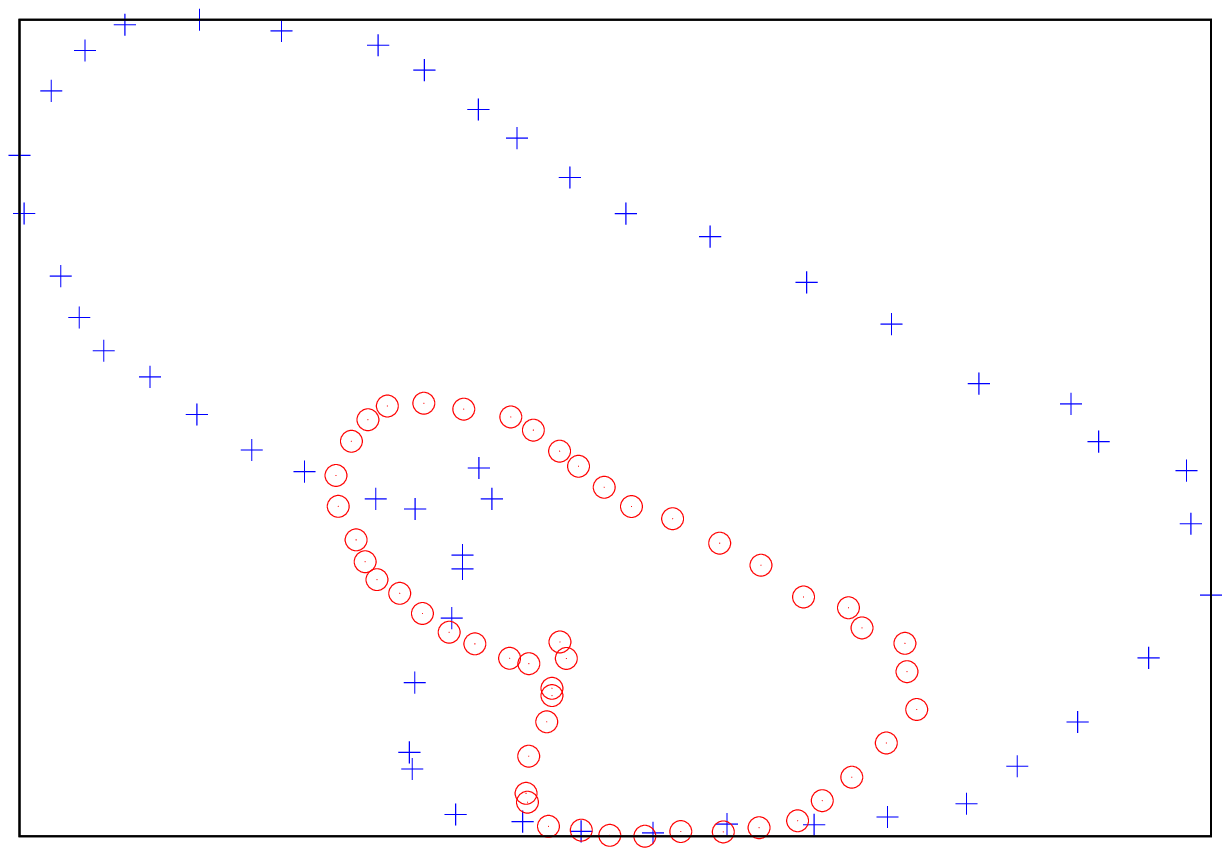}
 }
 \subfigure{
   \includegraphics [width=0.3\columnwidth]{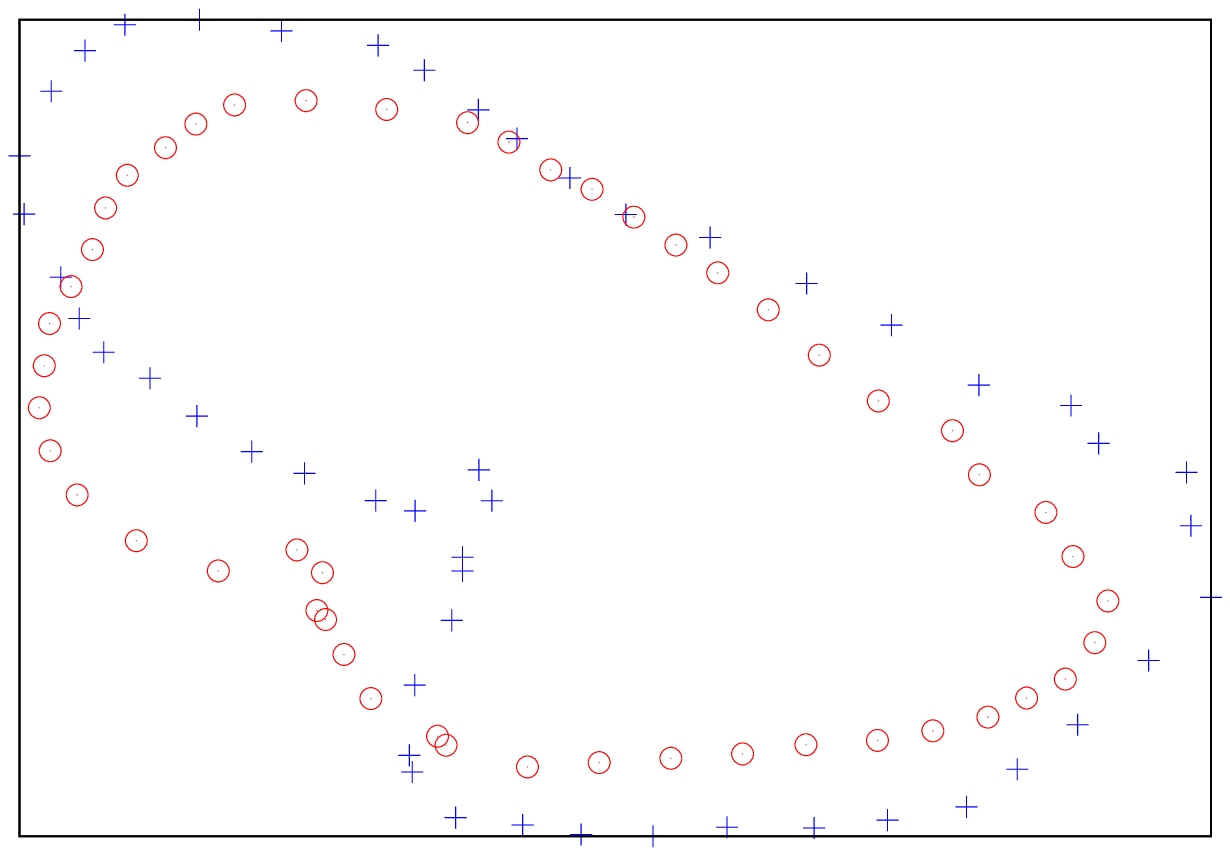}
 }\\{(b) point-set 2}\\
 \subfigure{
   \includegraphics [width=0.3\columnwidth]{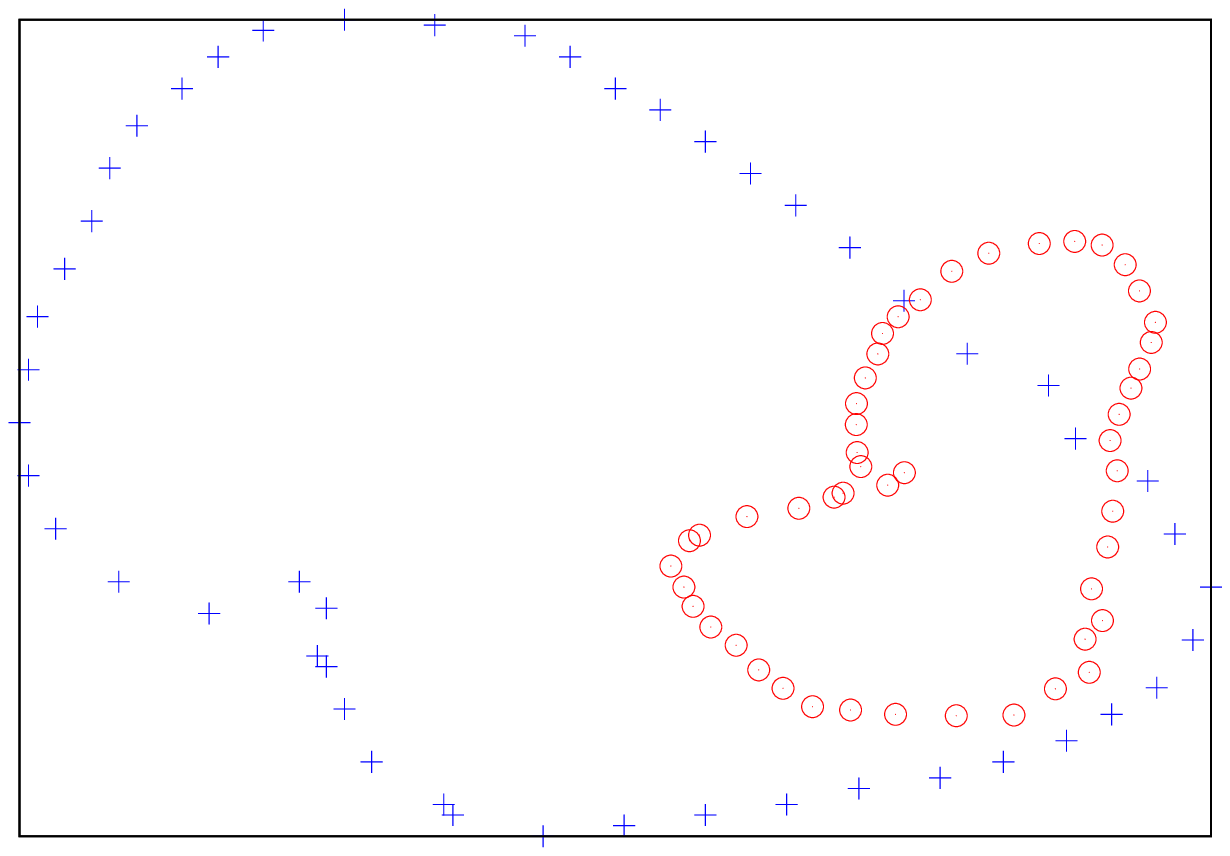}
 }
 \subfigure{
   \includegraphics [width=0.3\columnwidth]{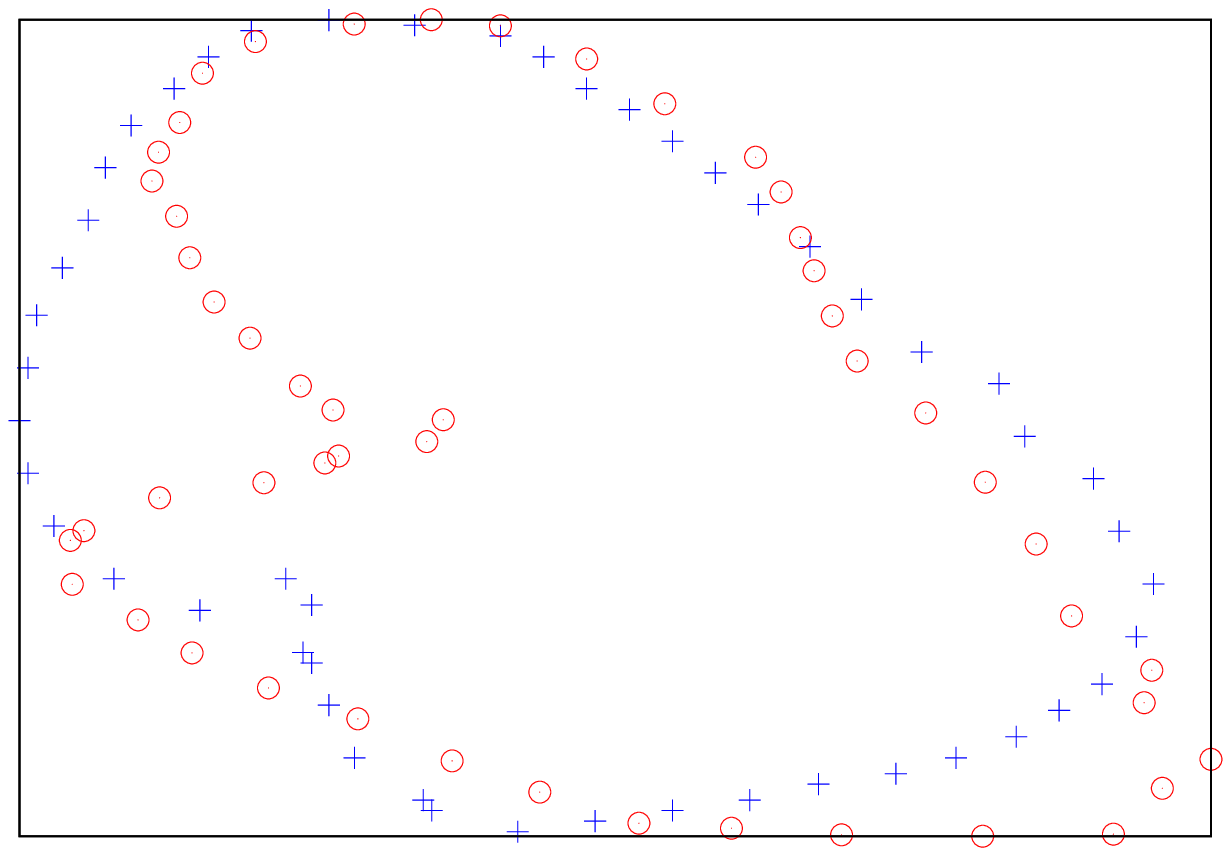}
 }\\{(c) point-set 3}\\
 \subfigure{
   \includegraphics [width=0.3\columnwidth]{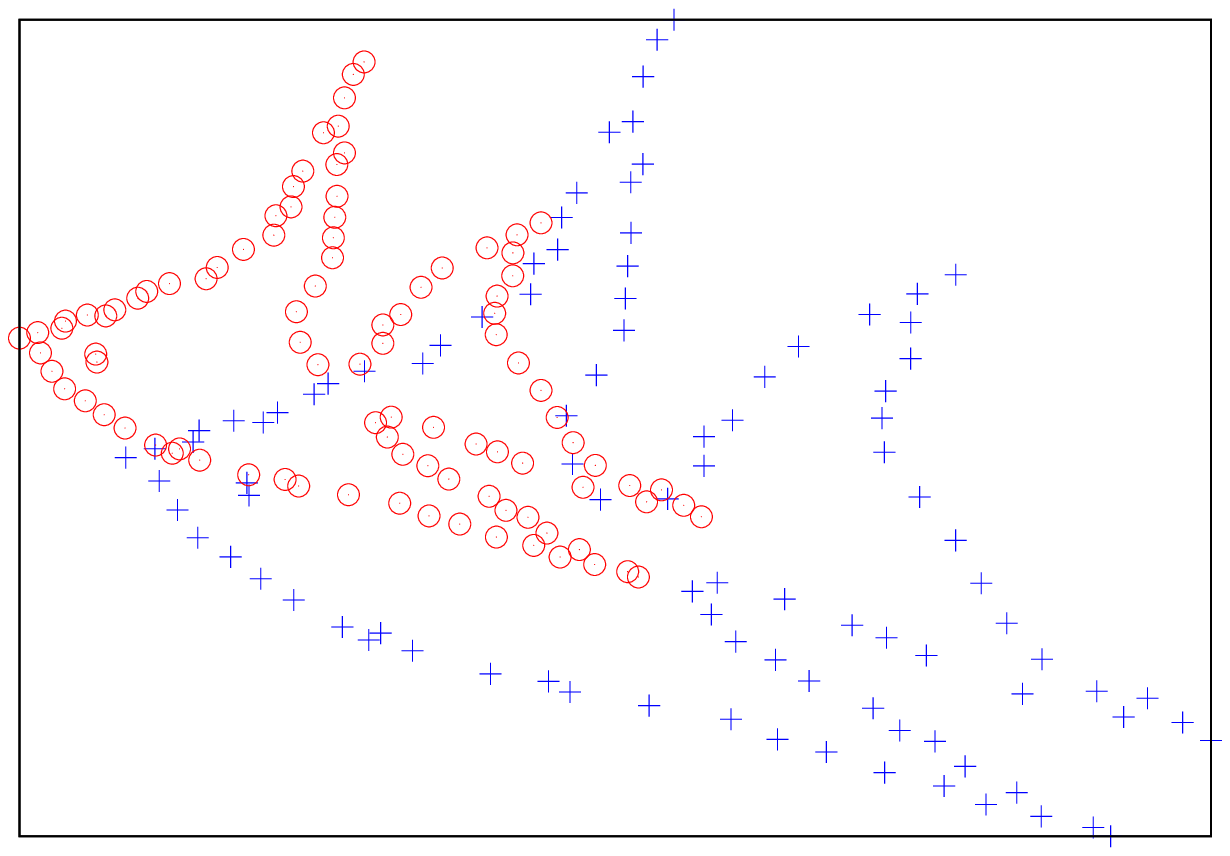}
 }
 \subfigure{
   \includegraphics [width=0.3\columnwidth]{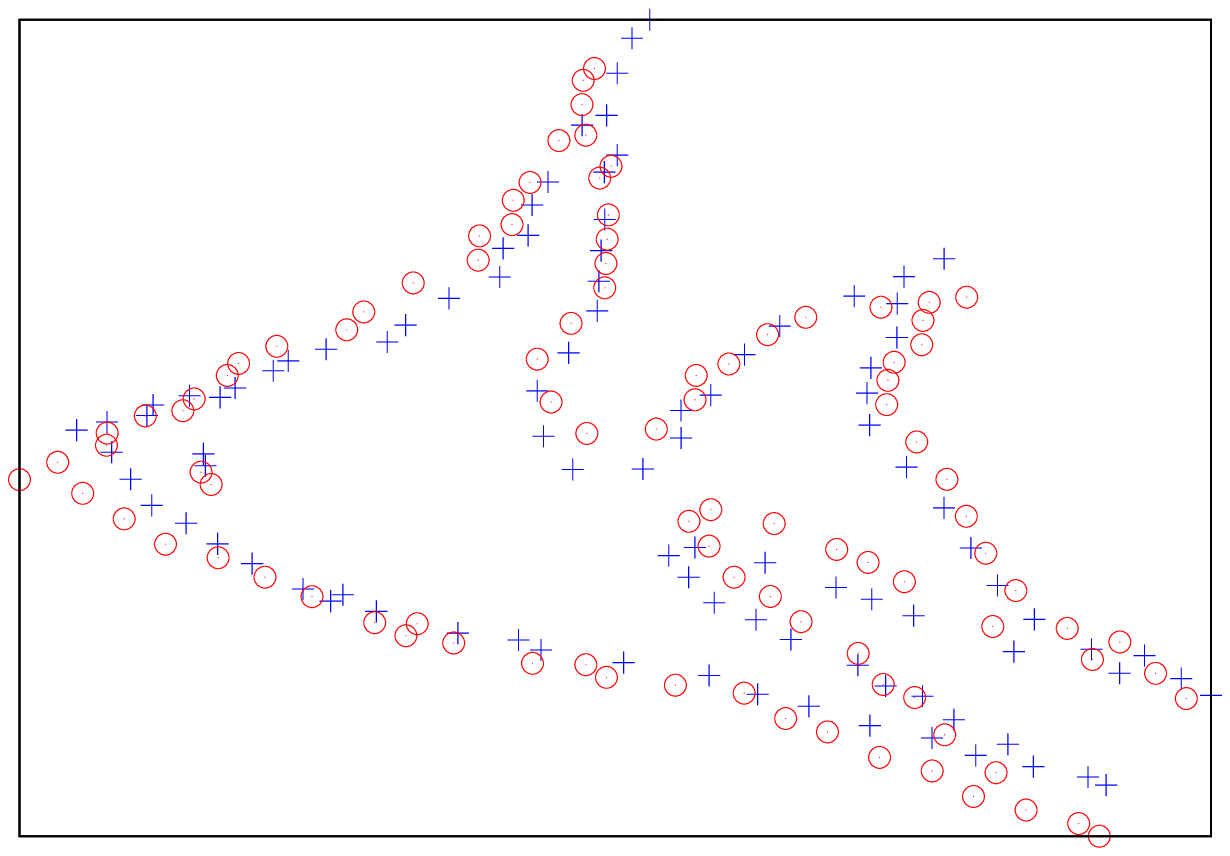}
 }\\{(d) point-set 4}\\
 \subfigure{
   \includegraphics [width=0.3\columnwidth]{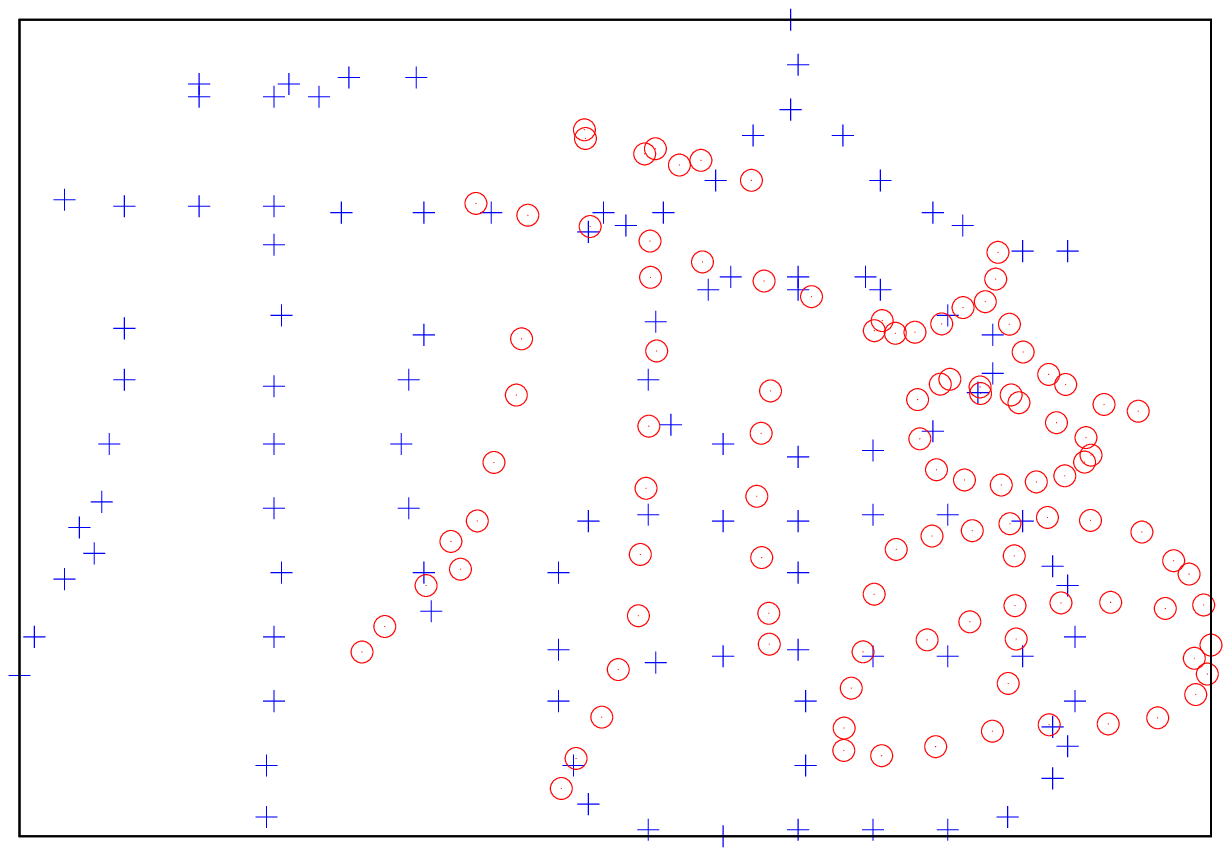}
 }
 \subfigure{
   \includegraphics [width=0.3\columnwidth]{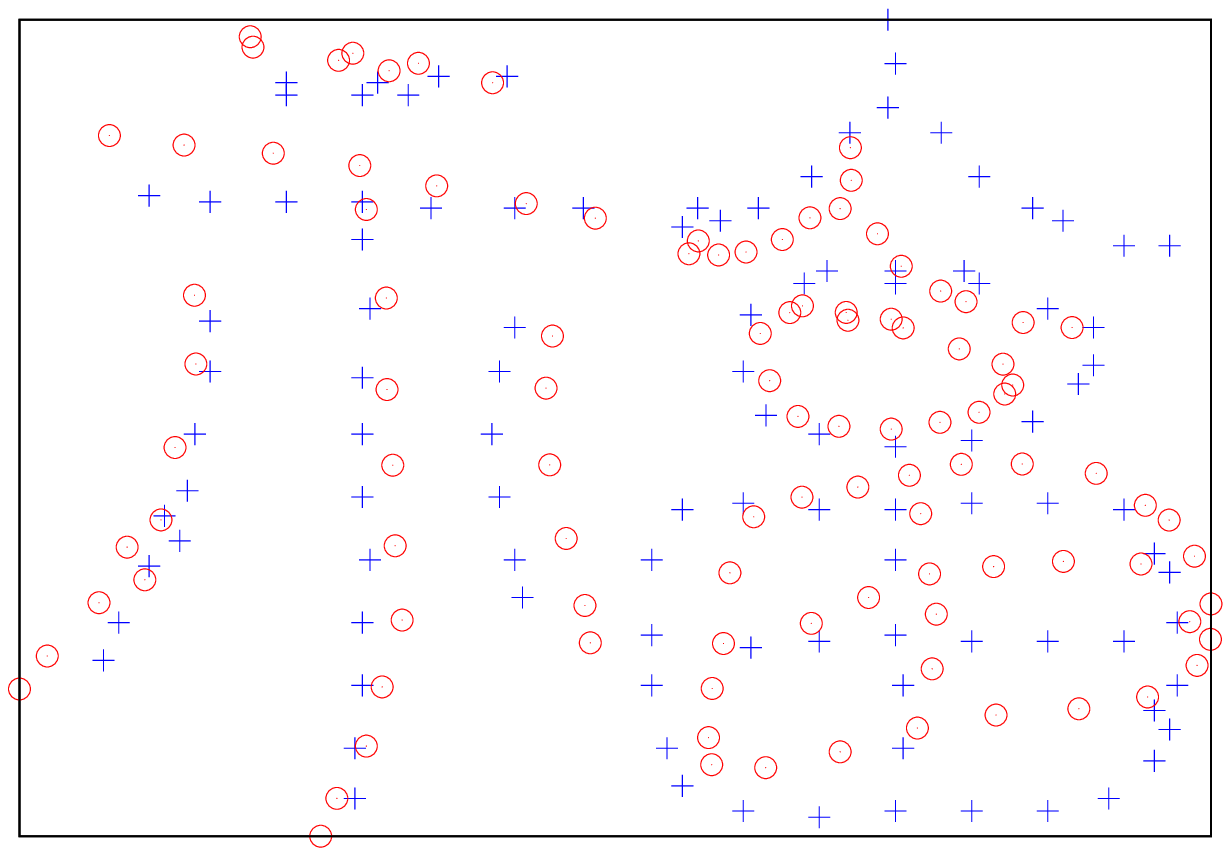}
 }\\{(e) point-set 5}\\
  \caption{Non-affine distorted point-sets (left, blue dots for static image points, red dots for deformed image points) and GA affine image registration (right, red dots are the warped image points) results obtained after 500 generations using population size 120. The warped images are zoomed for better visualization. Note that even better matching could be obtained with larger population sizes, but the improvements are negligible as shown in Figure~\ref{fig:non-affine}.} 
  \label{fig:non_affine_distorted}
\end{figure}

\begin{figure*}
  \centering
  \subfigure[point-set 1]{
      \includegraphics[width=0.31\textwidth]{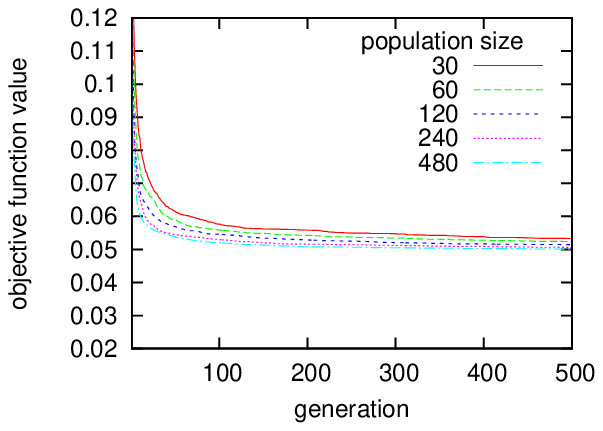}
  }
  \subfigure[point-set 2]{
      \includegraphics[width=0.31\textwidth]{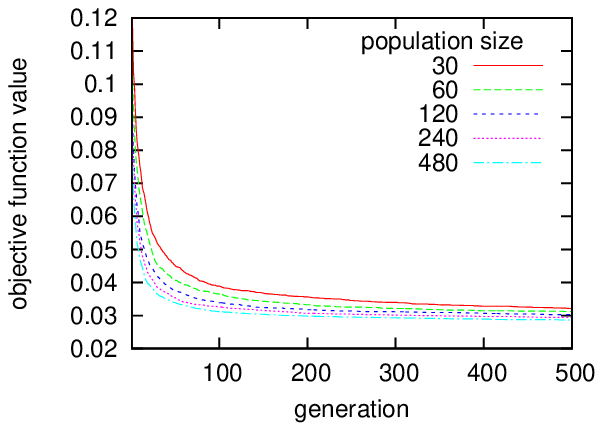}
  }
  \subfigure[point-set 3]{
      \includegraphics[width=0.31\textwidth]{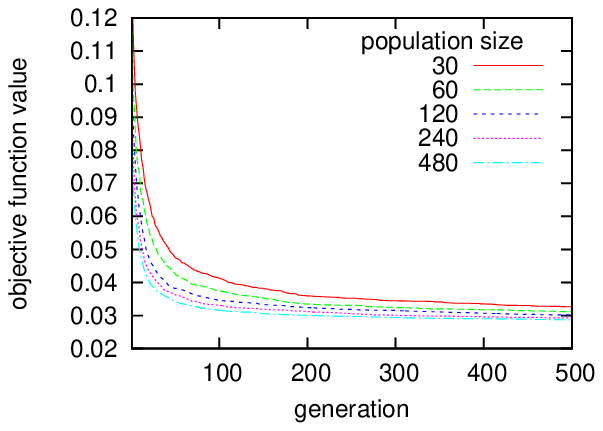}
  }
  \subfigure[point-set 4]{
      \includegraphics[width=0.31\textwidth]{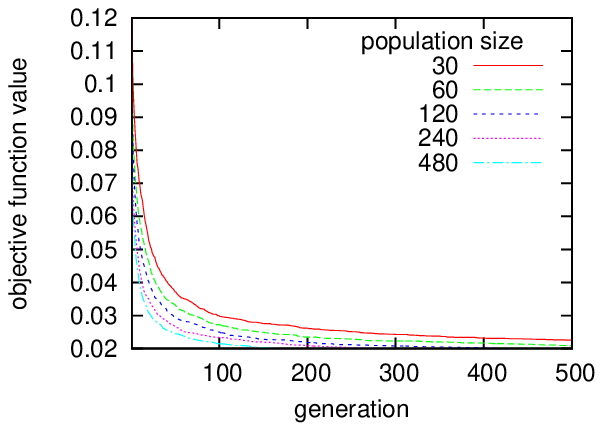}
  }
  \subfigure[point-set 5]{
      \includegraphics[width=0.31\textwidth]{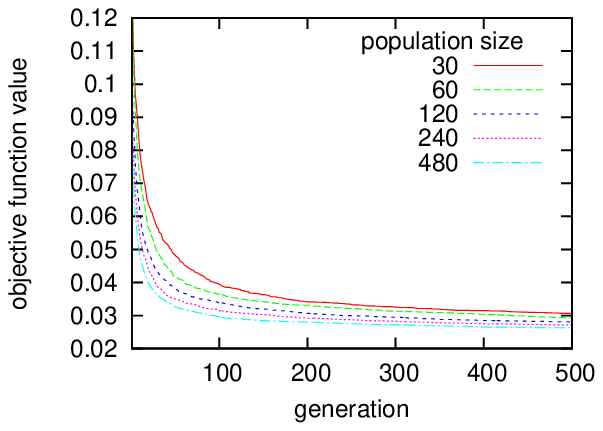}
  }
  \caption{Best objective function value through generations for various population sizes obtained for various point-sets. The results are averaged over 100 independent runs.} 
  \label{fig:non-affine}
\end{figure*}

\begin{figure*}
  \centering
  \subfigure{
   \includegraphics [width=3.7cm]{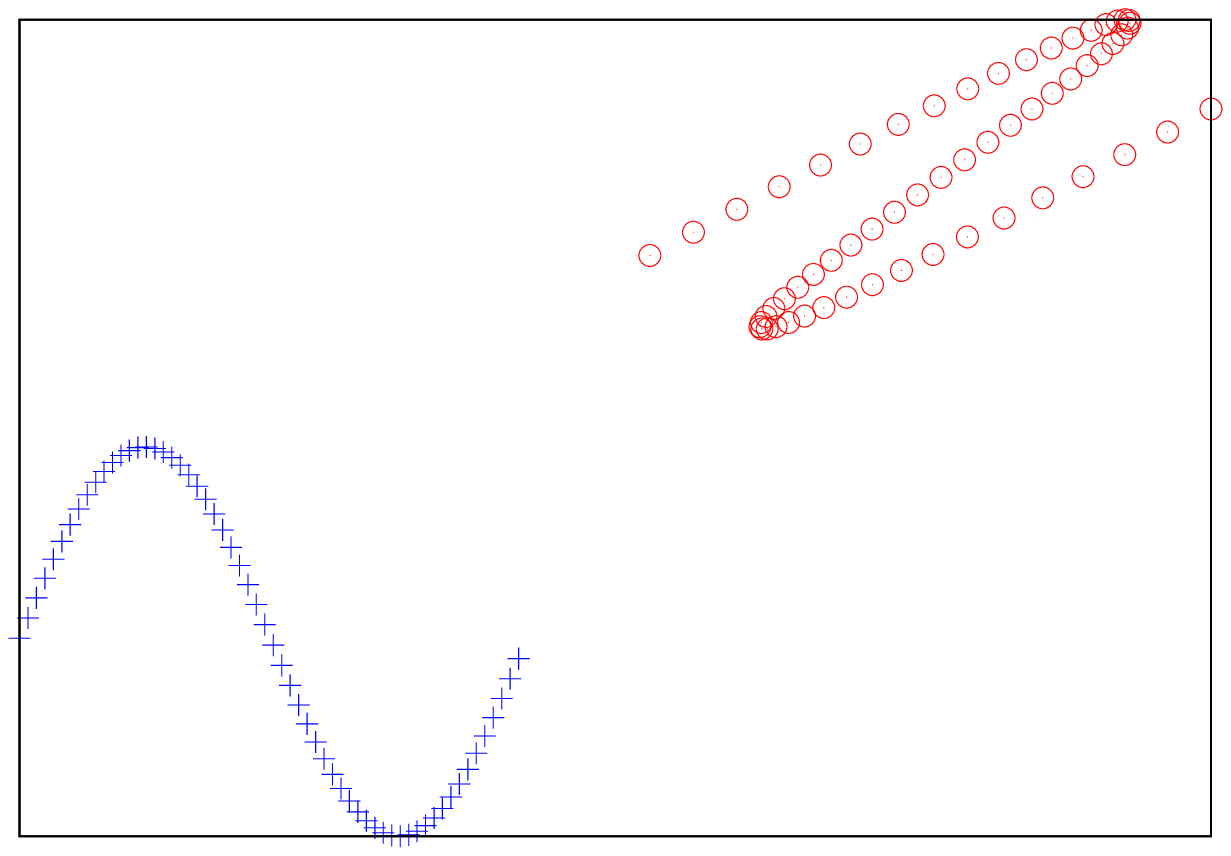}
 }
 \subfigure{
   \includegraphics [width=3.7cm]{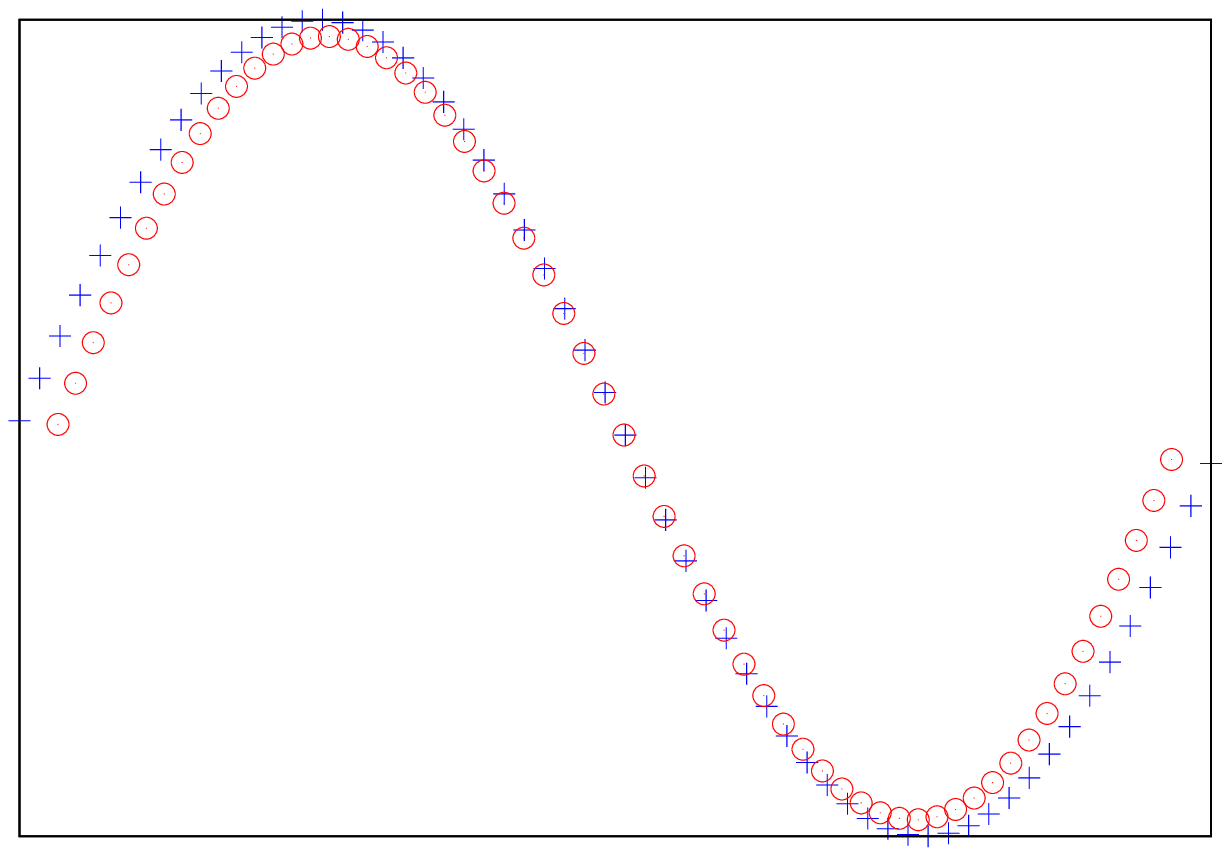}
 }
 \subfigure{
   \includegraphics [width=3.7cm]{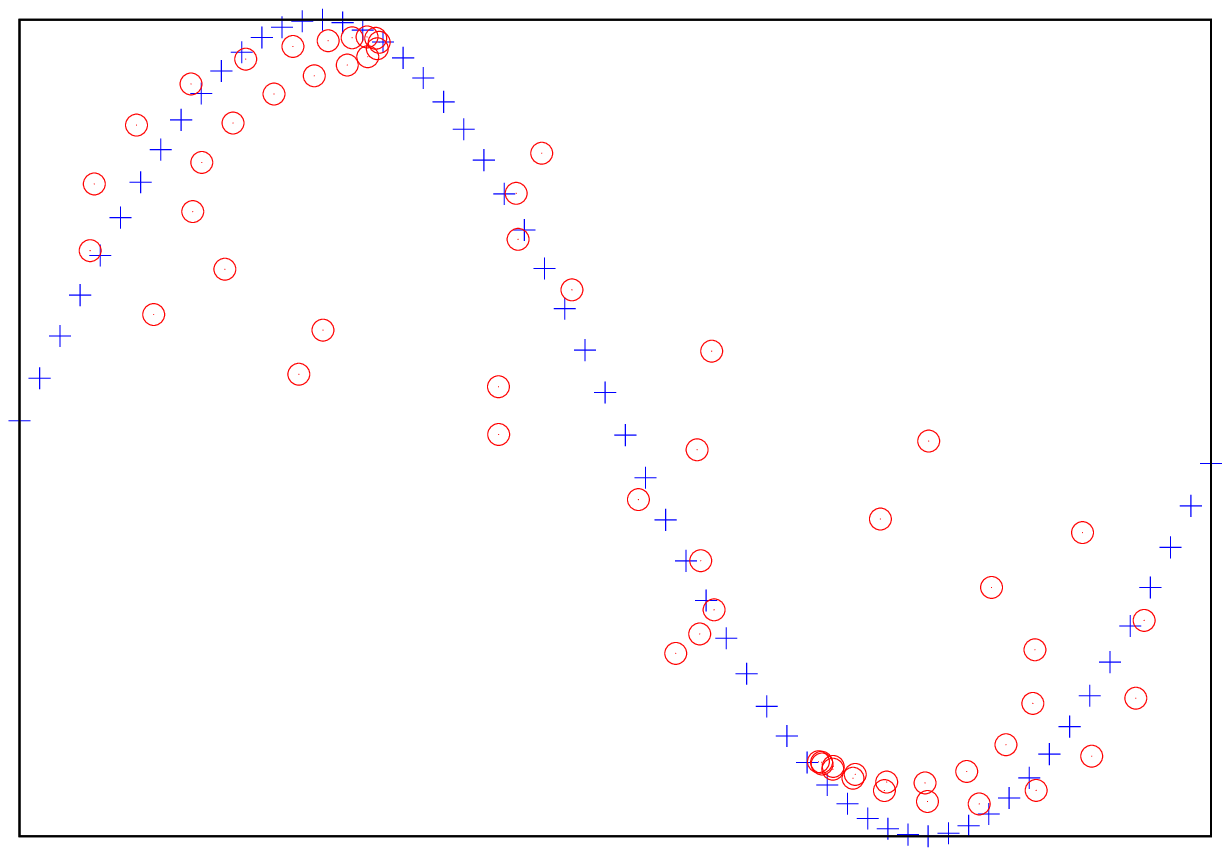}
 }
 \subfigure{
   \includegraphics [width=3.7cm]{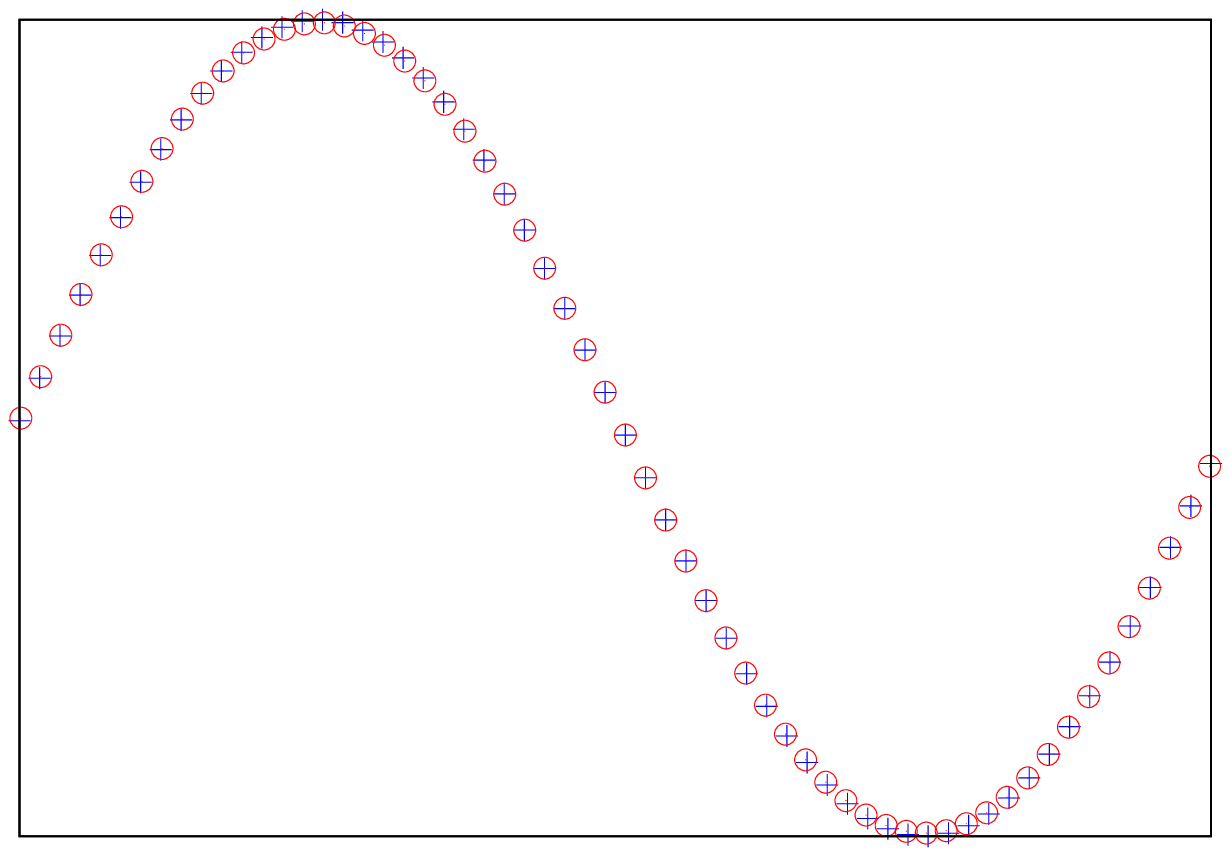}
 }\\
 \subfigure{
   \includegraphics [width=3.7cm]{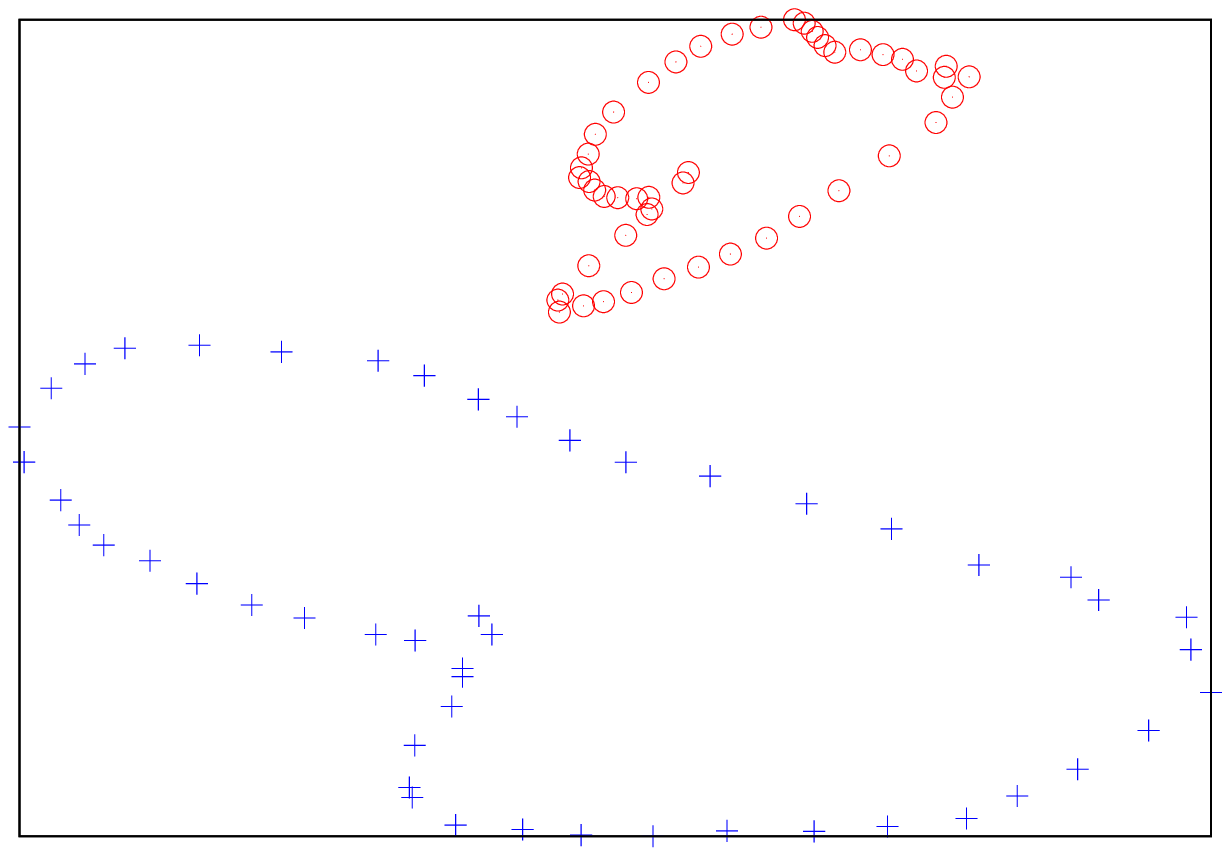}
 }
 \subfigure{
   \includegraphics [width=3.7cm]{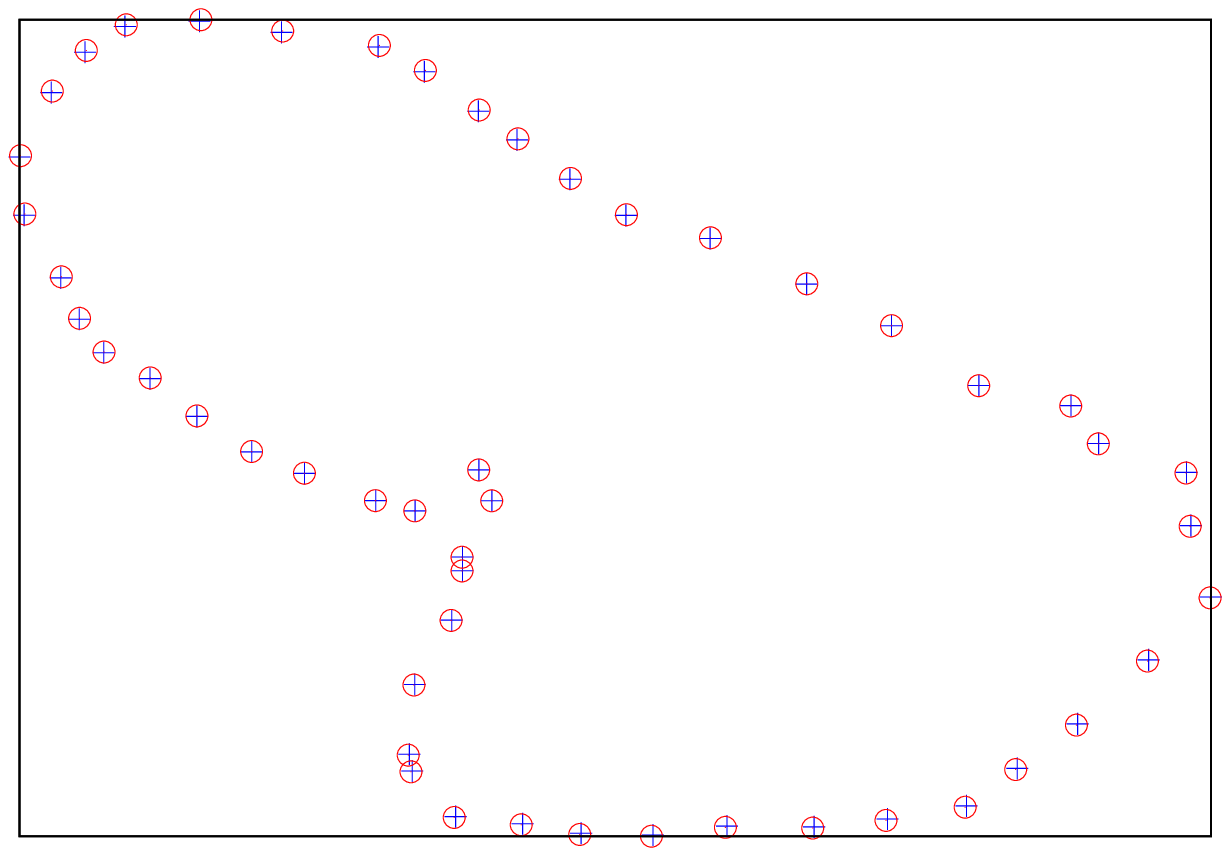}
 }
 \subfigure{
   \includegraphics [width=3.7cm]{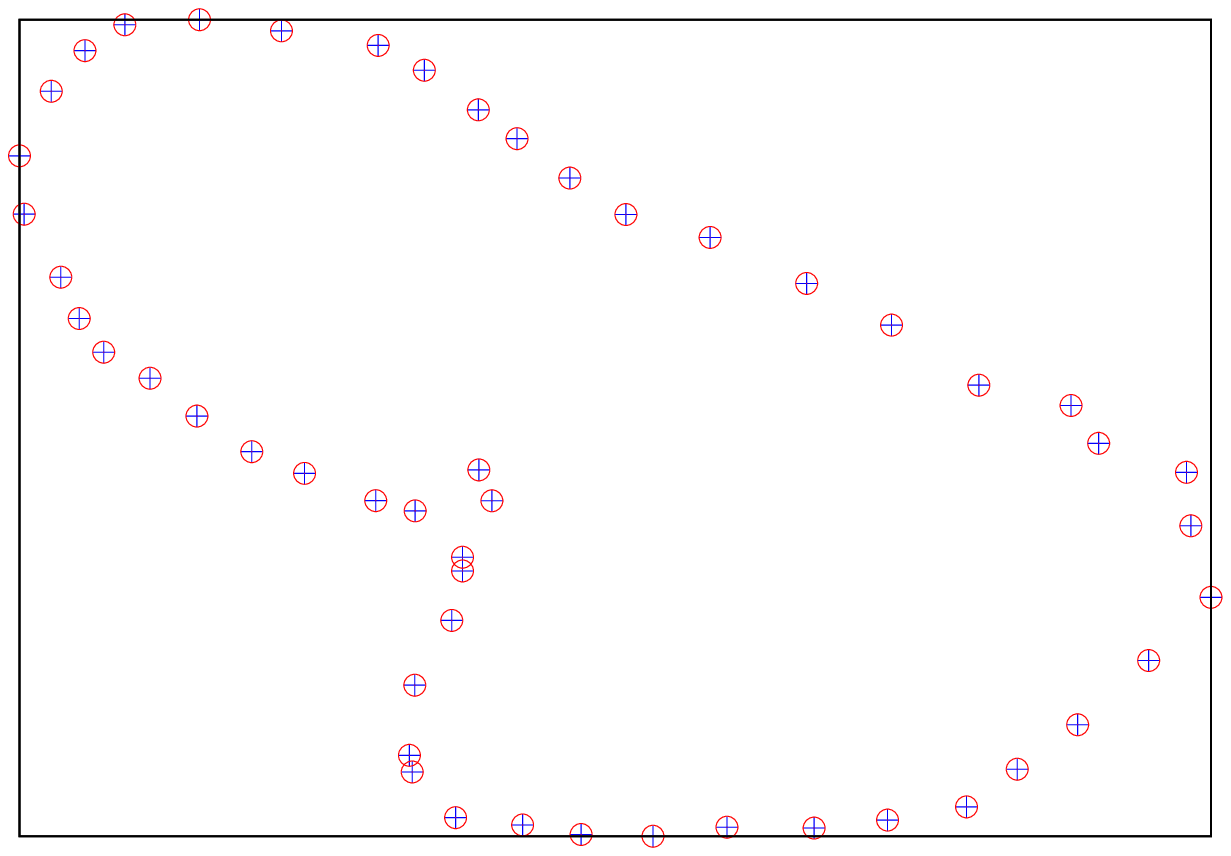}
 }
 \subfigure{
   \includegraphics [width=3.7cm]{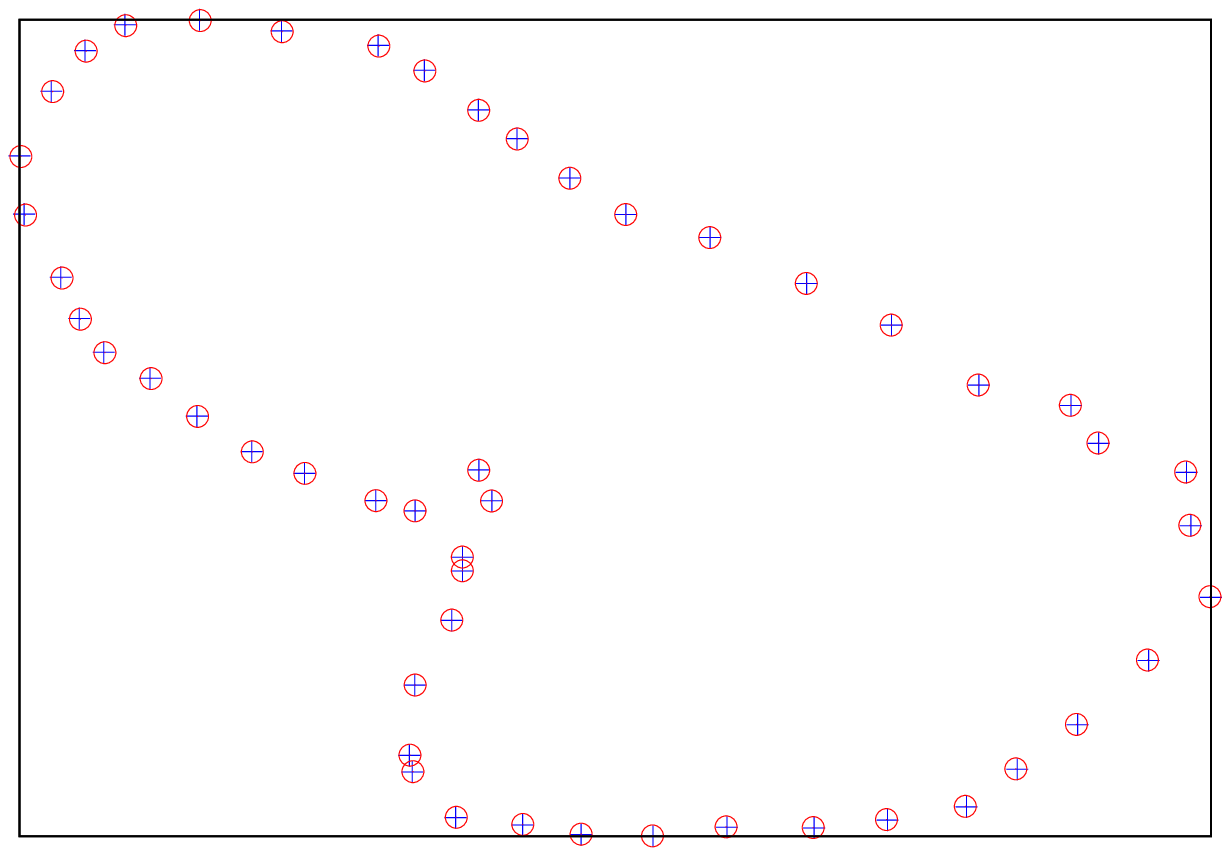}
 }\\
 \subfigure{
   \includegraphics [width=3.7cm]{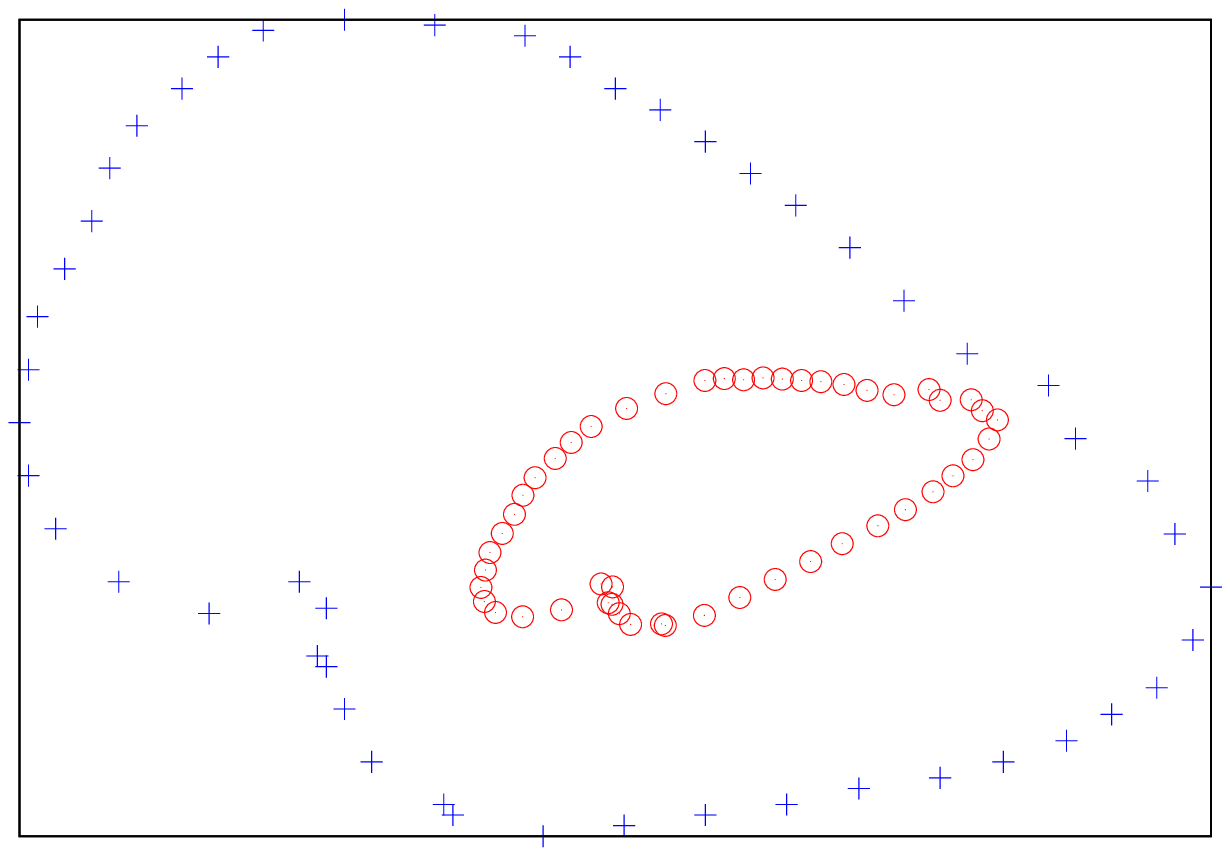}
 }
 \subfigure{
   \includegraphics [width=3.7cm]{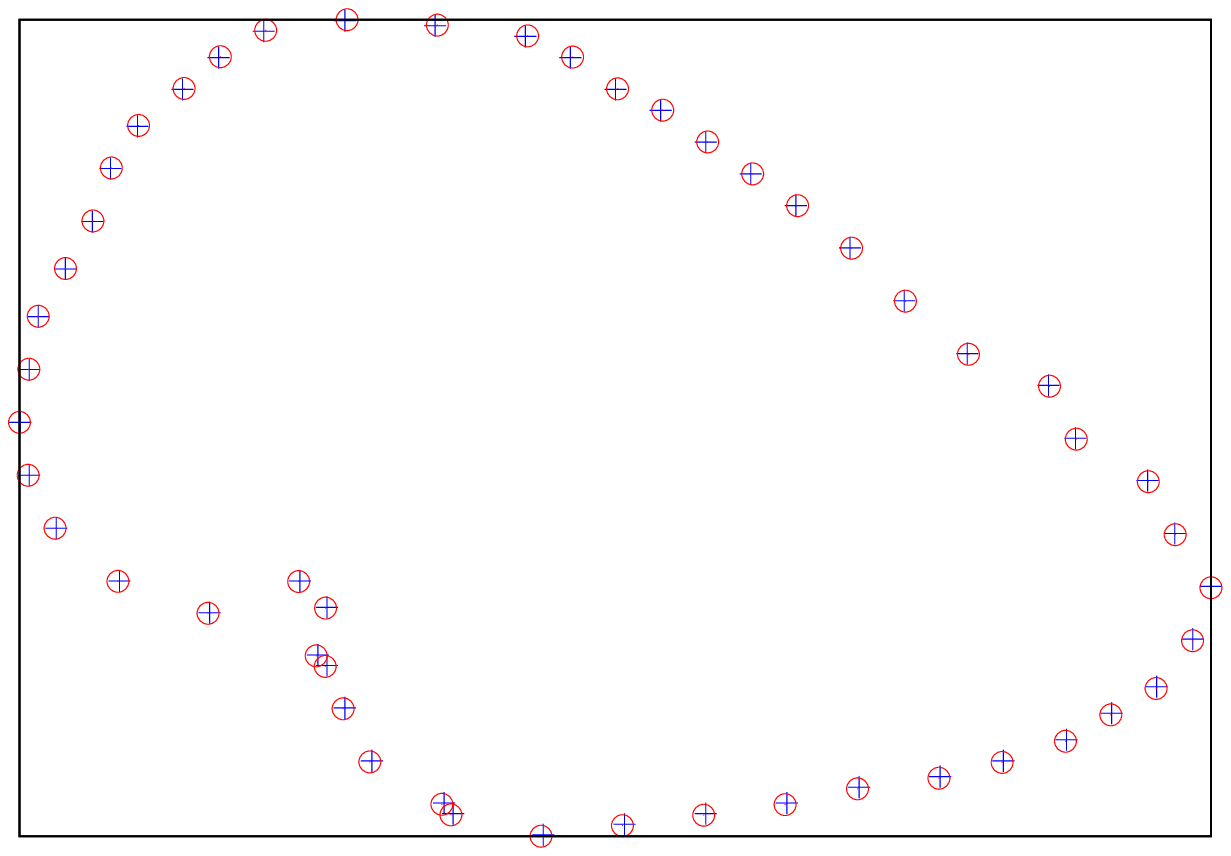}
 }
 \subfigure{
   \includegraphics [width=3.7cm]{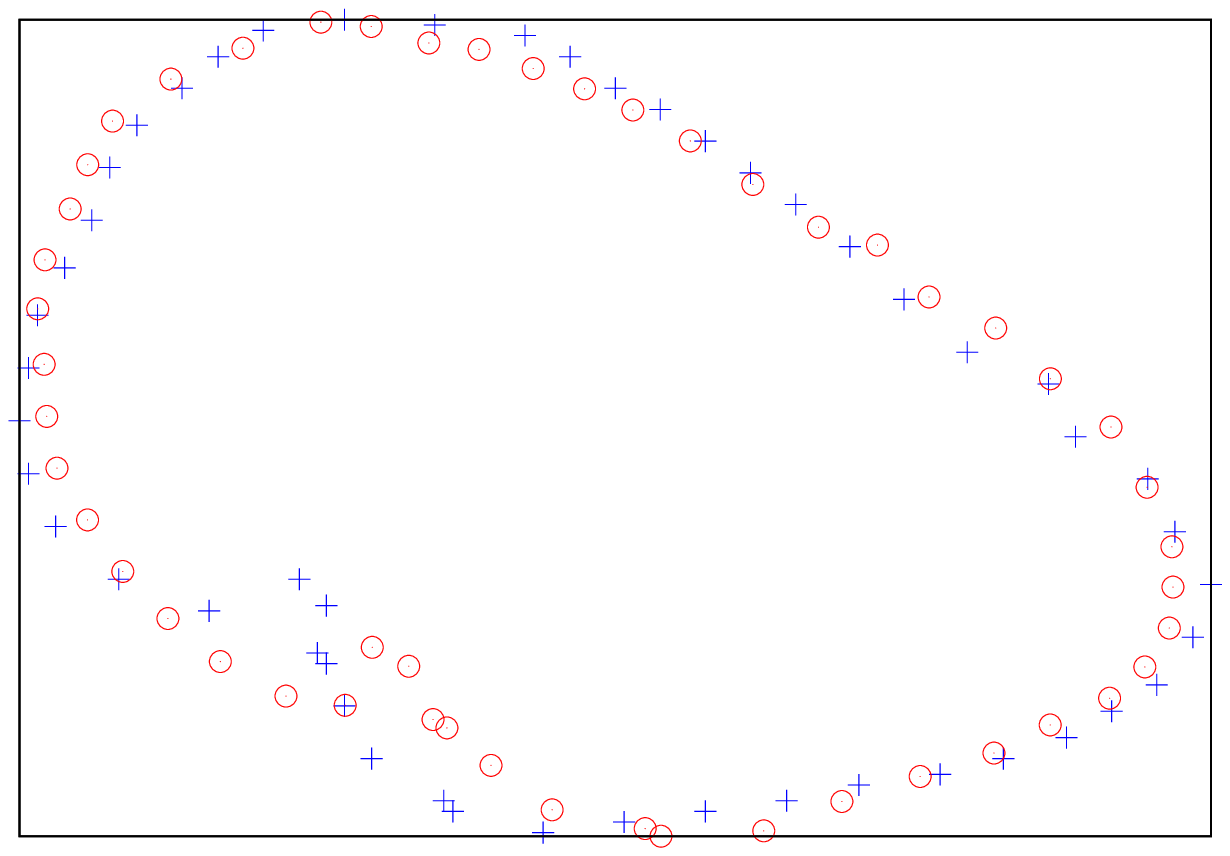}
 }
 \subfigure{
   \includegraphics [width=3.7cm]{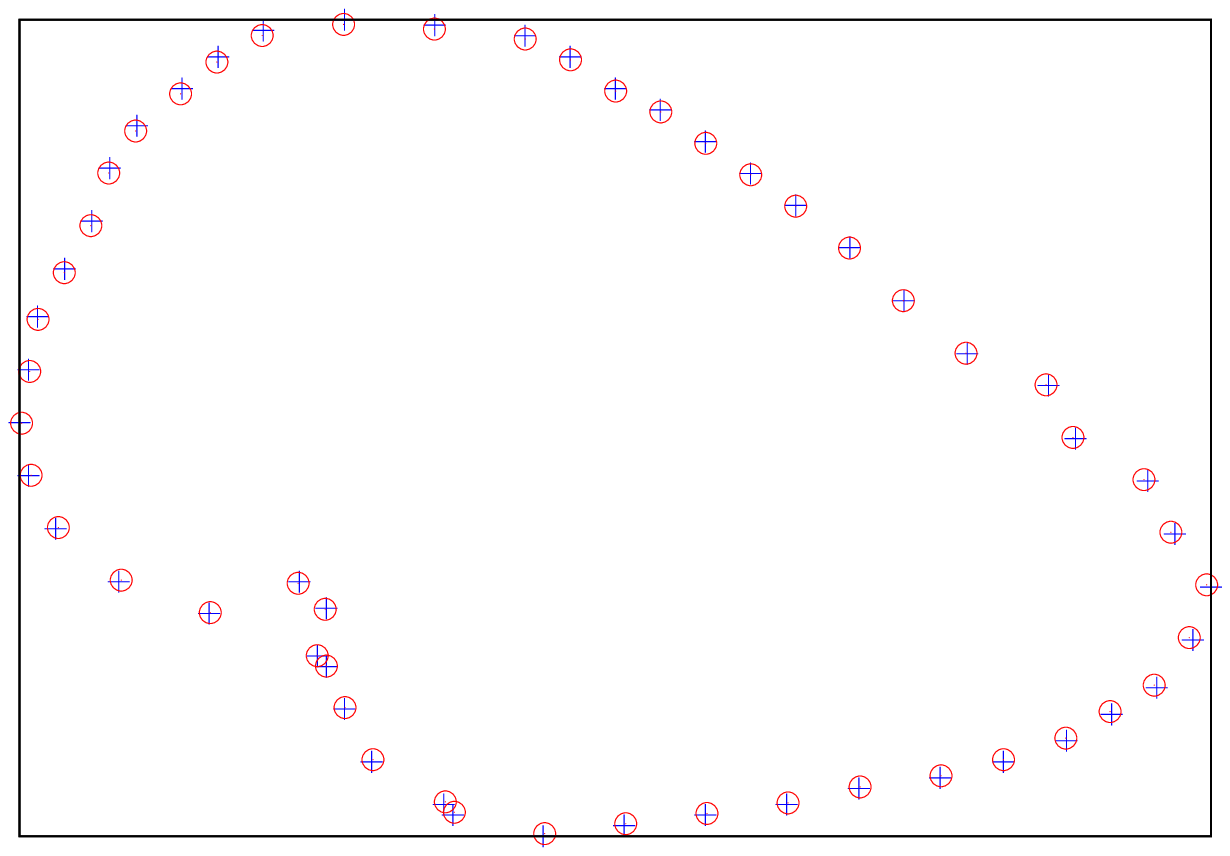}
 }\\
 \subfigure{
   \includegraphics [width=3.7cm]{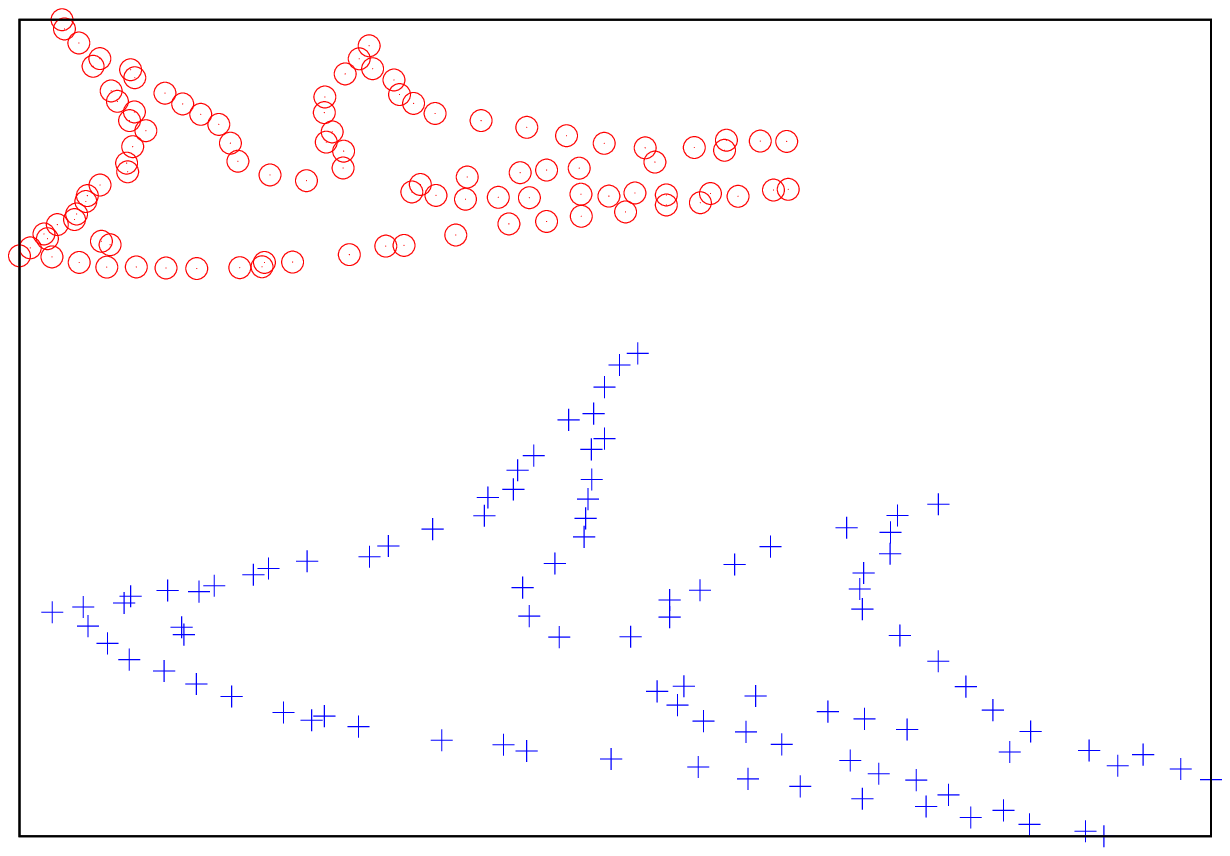}
 }
 \subfigure{
   \includegraphics [width=3.7cm]{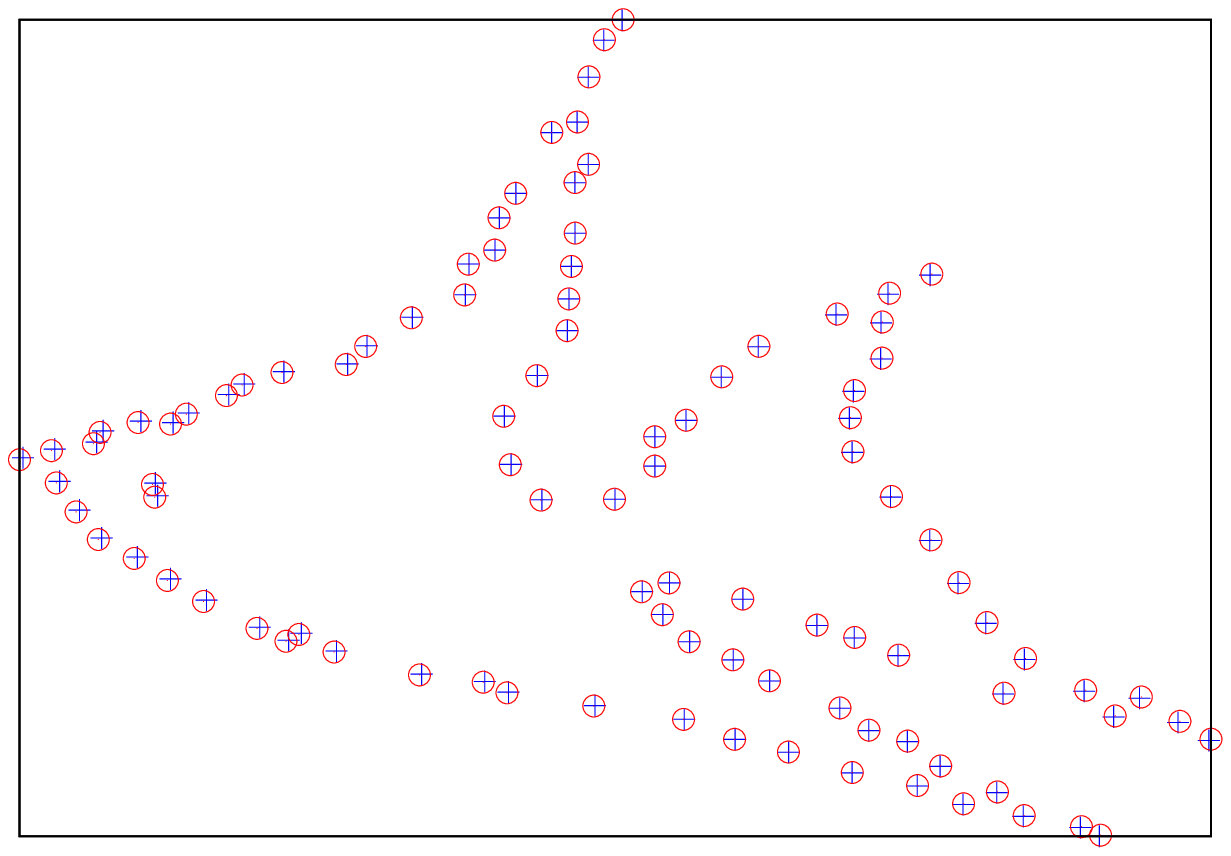}
 }
 \subfigure{
   \includegraphics [width=3.7cm]{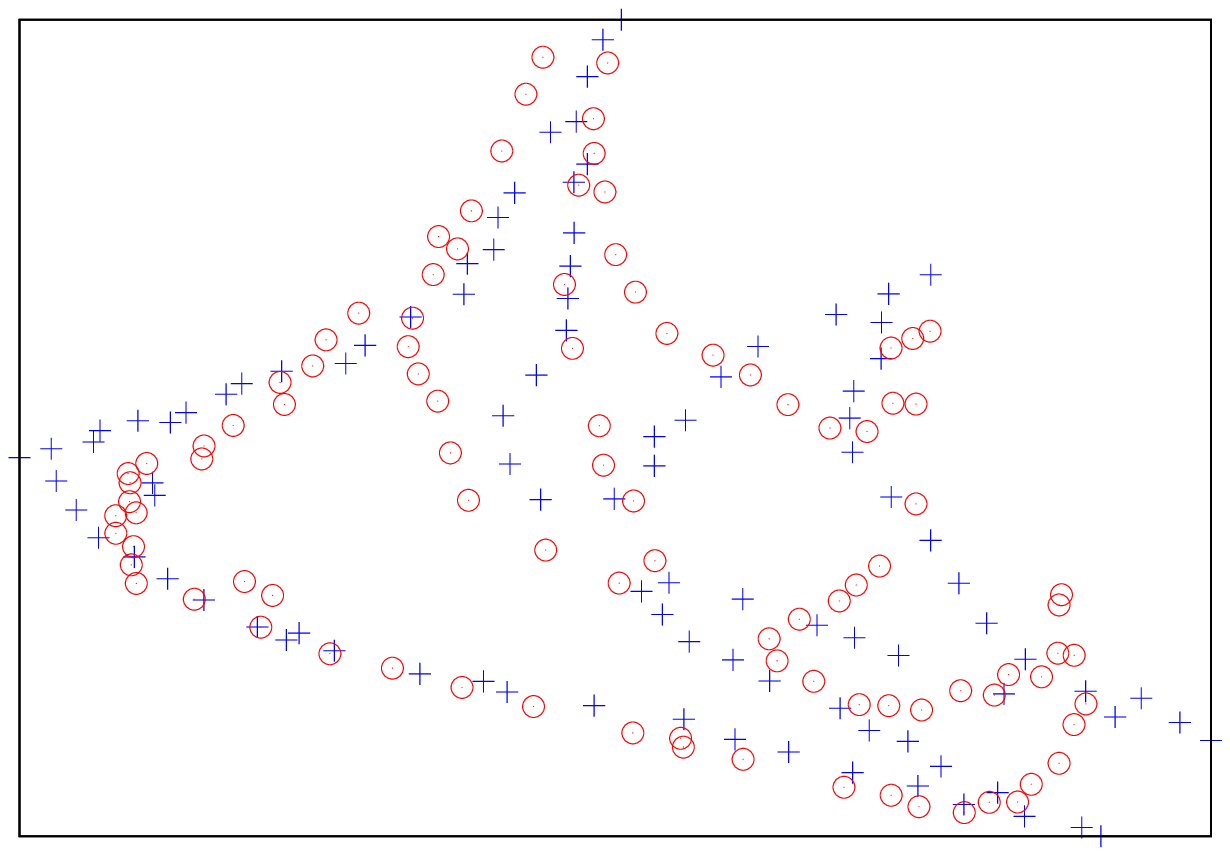}
 }
 \subfigure{
   \includegraphics [width=3.7cm]{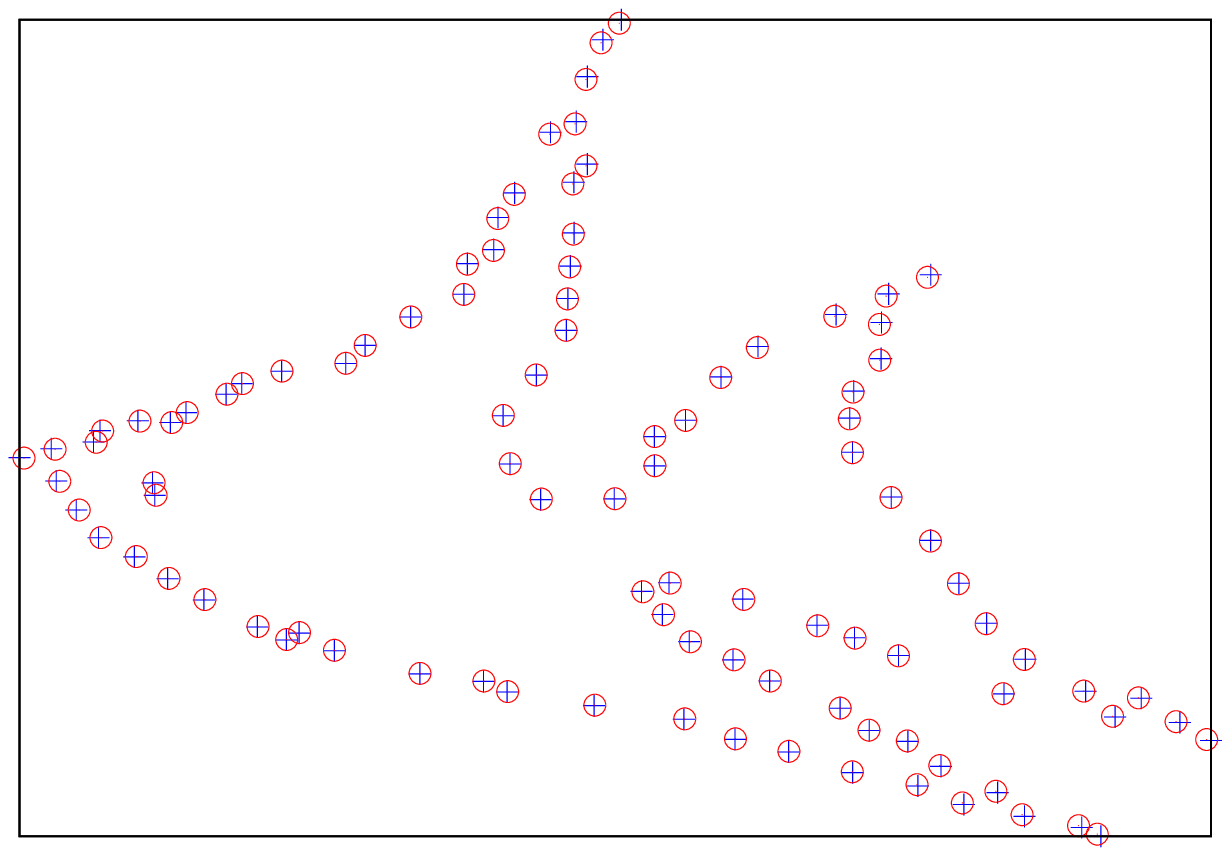}
 }\\
\setcounter{subfigure}{0} 
 \subfigure[point-sets]{
   \includegraphics [width=3.7cm]{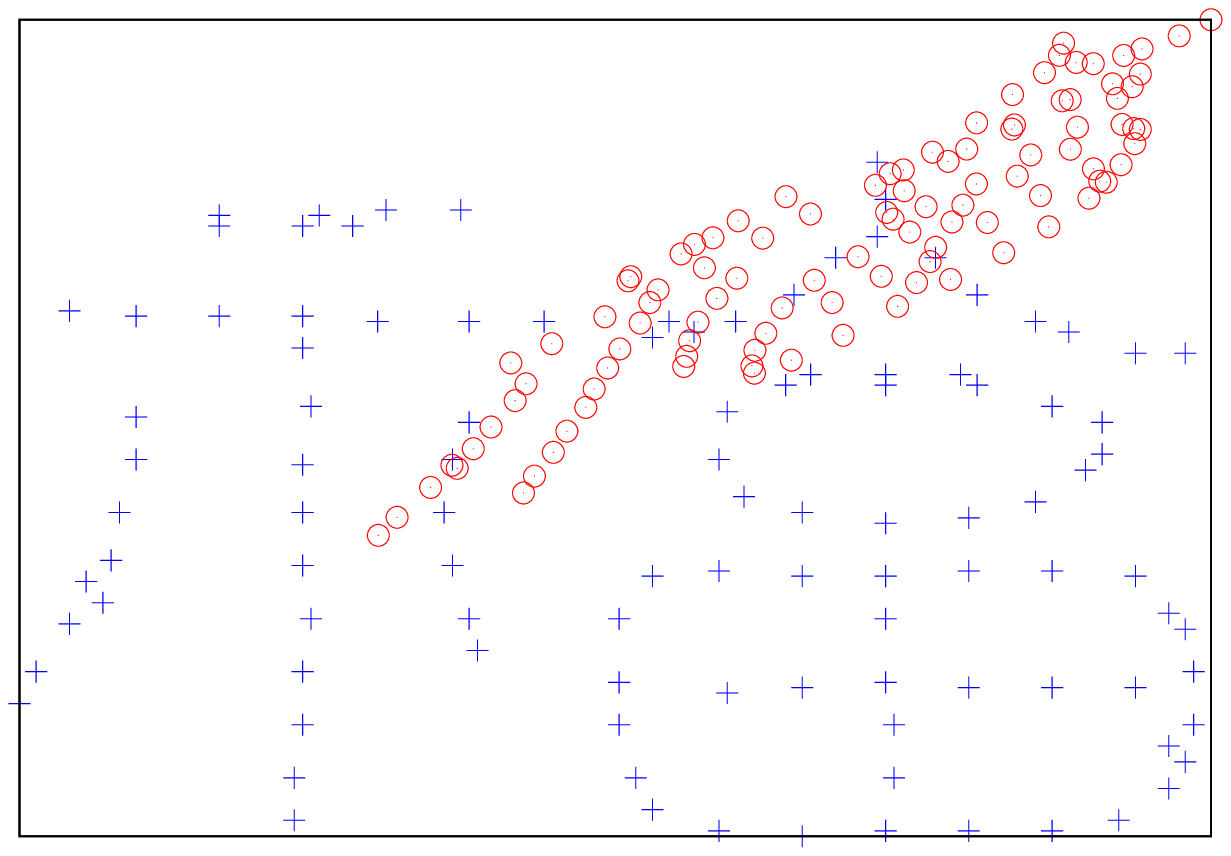}
   \label{subfig:affine_original_point_sets}
 }
 \subfigure[GA]{
   \includegraphics [width=3.7cm]{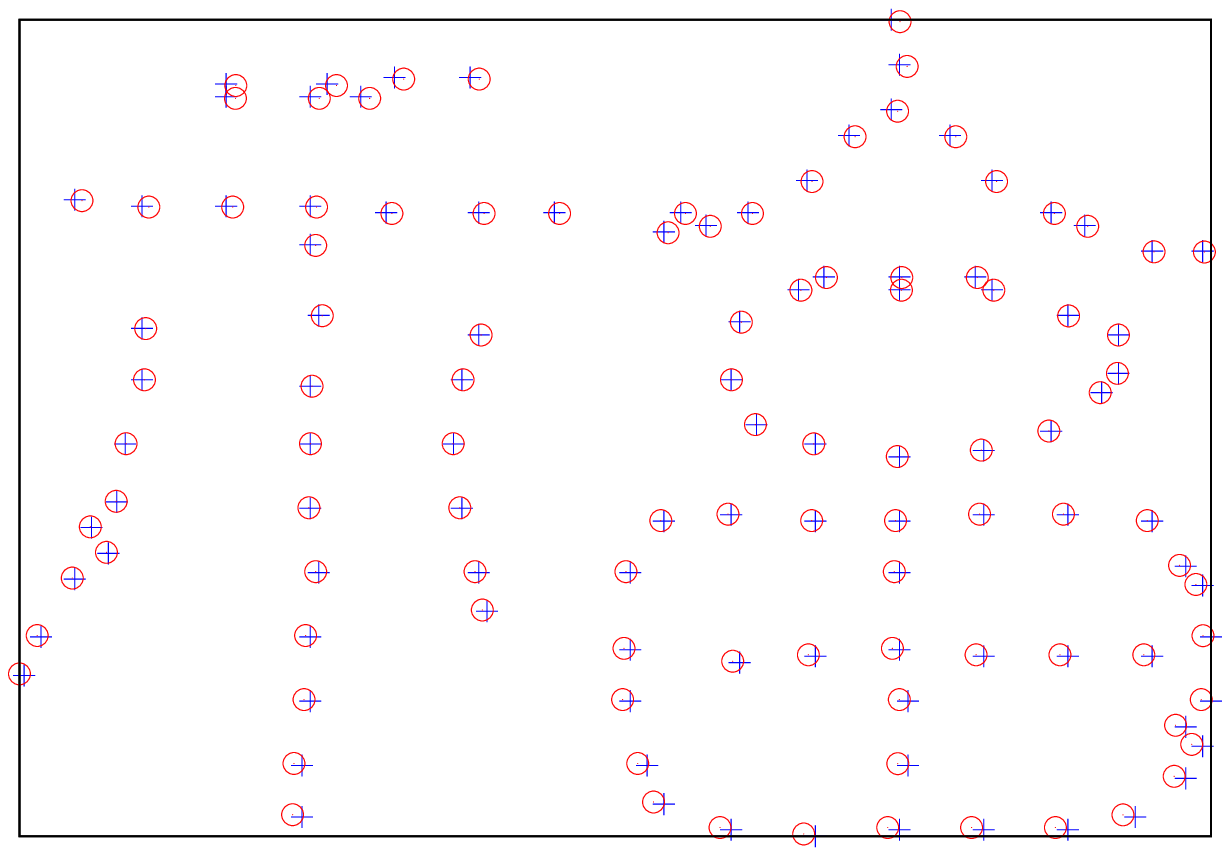}
   \label{subfig:affine_GA_point_sets}
 }
 \subfigure[SC]{
   \includegraphics [width=3.7cm]{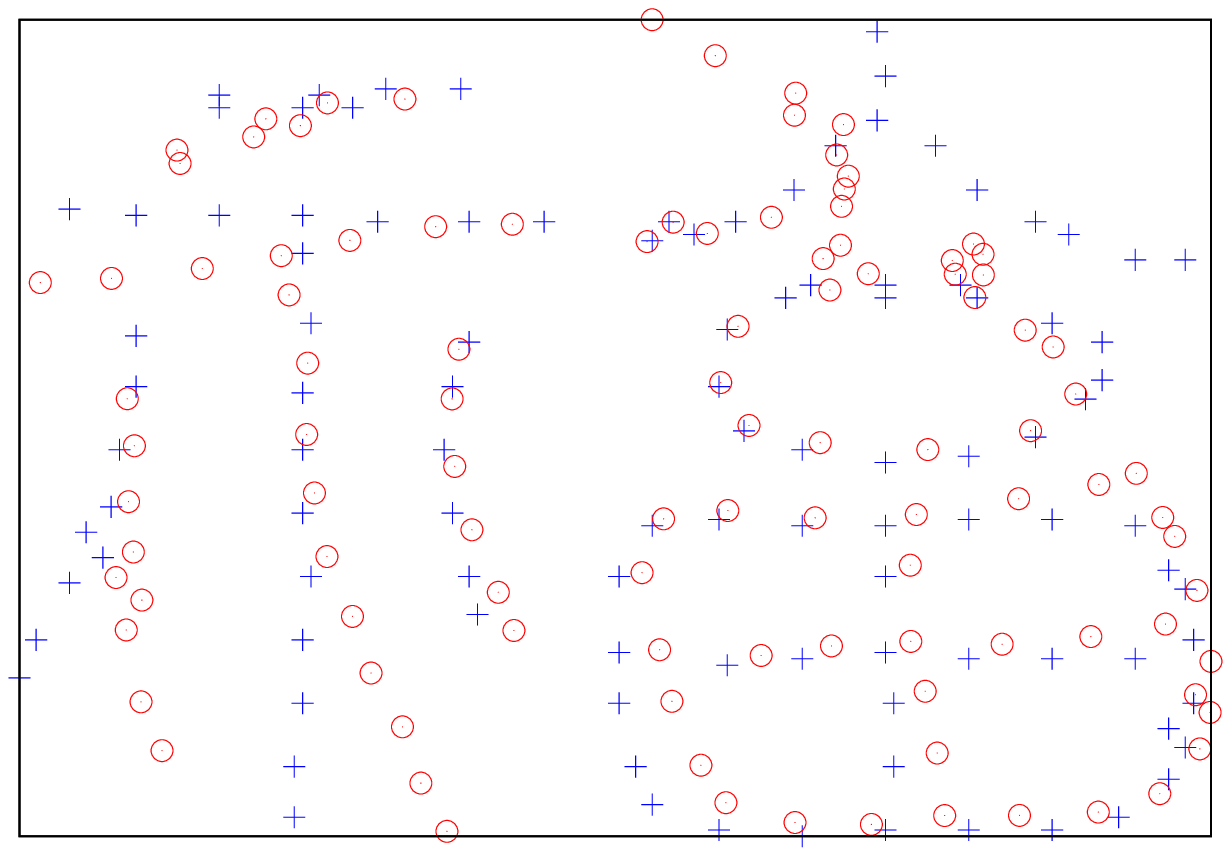}
   \label{subfig:affine_SC_point_sets}
 }
 \subfigure[TPS-RPM]{
   \includegraphics [width=3.7cm]{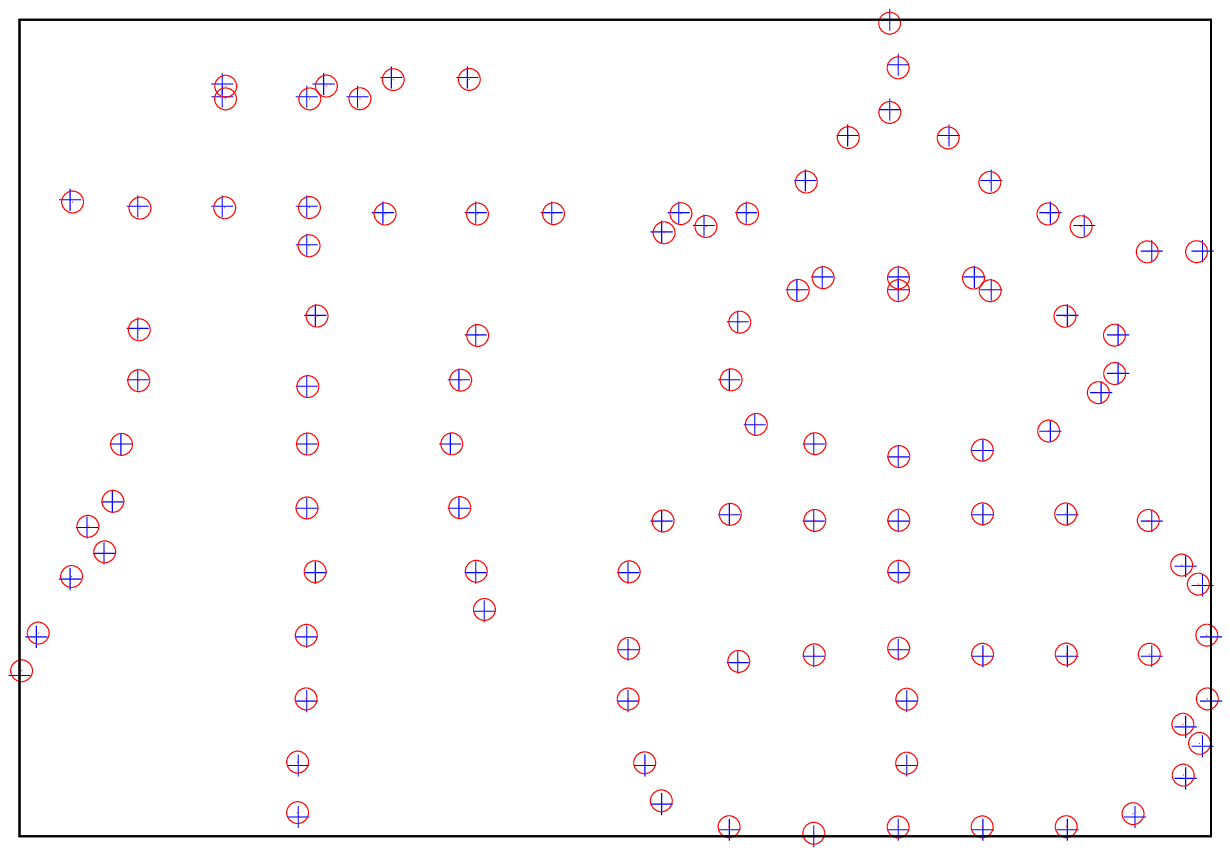}
   \label{subfig:affine_TPS_RPM_point_sets}
 }
  \caption{Affine distorted point-sets and respective registration results. On the left column, (a), the blue dots are the static image points and the red dots are the deformed image points. On the other columns, (b), (c), and (d), the red dots are the warped image points.
The warped images are zoomed for better visualization.
The GA results were obtained after 500 generations using population size 120. Observe that for the case of GA and TPS-RPM, the deformed and static points are almost on top of each other, meaning that the match is almost perfect. For SC the results are slightly inferior compared with those obtained by the GA and TPS-RPM.} 
  \label{fig:affine_distorted}
\end{figure*}

Figure~\ref{fig:non_affine_distorted} illustrates the five point-sets before and after warping. It should be noticed that these are sets where deformations are non-affine, therefore the resulting warped images will never match perfectly. Nevertheless, they present a very good approximation. When compared to the approach from~\cite{SeixasOchiConciSaade:2008} it is possible to observe that our results are more precise. 

In order to assess the quality of the proposed approach in what concerns affine deformations, affine deformed images were generated from the same five static images, and used in subsequent experiments. The results were compared to those produced by well known classical state-of-the-art approaches such as SC and TPS-RPM, and are illustrated in Figure~\ref{fig:affine_distorted}. It can be seen that the results produced by the proposed GA are slightly better than those of the SC and are very competitive with those obtained by TPS-RPM. 

~\\

~\\

\section{Summary and Conclusions}
\label{sec:conclusion}
This paper proposed a real coded GA that is especially suited for doing image registration of affine distorted images. As opposed to previous EC approaches for solving IR, our method uses Simulated Binary Crossover, a parent-centric recombination operator that has been giving good results on a variety of continuous real world optimization problems within a GA framework. The use of a randomized ordering when visiting points during the point-matching procedure was also proposed, and although this technique yields a noisy fitness function evaluation, the results obtained show that the GA is capable of dealing with it quite well. 

The resulting algorithm was applied to 2-D synthetic point-sets, with deformed images points obtained from both affine and non-affine transformations. For the case of non-affine distorted points, our method produces a more precise registration than previously published results by means of an evolutionary algorithm on the same point-sets~\cite{SeixasOchiConciSaade:2008}. For the case of affine distorted points, the proposed real coded GA produced better results than SC, and was competitive with TPS-RPM, two well known classical state-of-the-art image registration methods.

\section{Acknowledgements}
The authors would like to thank Faroq AL-Tam for valuable comments.

\bibliographystyle{plain}
\bibliography{IR-sGA}

\end{document}